\documentclass{article} 
\usepackage{iclr2025_conference,times}

\usepackage{amsmath,amsfonts,bm}

\def\eqref#1{equation~\ref{#1}}

\def\1{\bm{1}}

\DeclareMathAlphabet{\mathsfit}{\encodingdefault}{\sfdefault}{m}{sl}
\SetMathAlphabet{\mathsfit}{bold}{\encodingdefault}{\sfdefault}{bx}{n}

\usepackage{hyperref}
\usepackage{url}

\usepackage{graphicx}
\usepackage{subfigure}
\usepackage{amsmath}
\usepackage{multirow}
\usepackage{listings}
\usepackage{array}
\usepackage{xspace}
\usepackage{booktabs}
\usepackage{colortbl}
\definecolor{LightCyan}{rgb}{0.8,1,1}
\usepackage{pifont}

\usepackage{tikz}
\usepackage[most]{tcolorbox}

\newcommand{\modelname}{\textsc{PathSeeker}\xspace}

\newcommand{\benchname}{BioMaze\xspace}

\definecolor{ForestGreen}{RGB}{34,139,34}

\newcommand{\finished}[1]{\textcolor{red}{}}
\lstdefinestyle{QuestionType}{
  basicstyle=\ttfamily\color{blue},
  stringstyle=\ttfamily\color{blue},
  breaklines=true,
  breakatwhitespace=true,
  showspaces=false,
  showstringspaces=false,
  breakindent=0pt,
}

\lstdefinestyle{AnswerType}{
  basicstyle=\ttfamily\color{ForestGreen},
  stringstyle=\ttfamily\color{ForestGreen},
  breaklines=true,
  breakatwhitespace=true,
  showspaces=false,
  showstringspaces=false,
  breakindent=0pt,
}

\lstdefinestyle{ModelResponseType}{
  basicstyle=\ttfamily\color{red},
  stringstyle=\ttfamily\color{red},
  breaklines=true,
  breakatwhitespace=true,
  showspaces=false,
  showstringspaces=false,
  breakindent=0pt,
}

\lstdefinestyle{JudgeType}{
  basicstyle=\ttfamily\color{purple},
  stringstyle=\ttfamily\color{purple},
  breaklines=true,
  breakatwhitespace=true,
  showspaces=false,
  showstringspaces=false,
  breakindent=0pt,
}

\title{\benchname: Benchmarking and Enhancing Large Language Models for Biological Pathway Reasoning}

\author{Haiteng Zhao$^1$\hspace{0.3cm} Chang Ma$^2$ \hspace{0.3cm} Fangzhi Xu$^3$ \hspace{0.3cm}  \textbf{Lingpeng Kong$^2$} \hspace{0.3cm} \textbf{Zhi-Hong Deng$^1$} \\
$^1$ Peking University $^2$ \hspace{0.8cm} The University of Hong Kong 
\hspace{0.8cm} $^3$ Xi'an Jiaotong University \\
\texttt{\{zhaohaiteng, zhdeng\}@pku.edu.cn \hspace{0.0cm}} \\
\texttt{\{cma, lpk\}@cs.hku.hk \hspace{0.0cm} fangzhixu98@gmail.com} \\
}

\iclrfinalcopy 
\begin{document}

\maketitle

\begin{abstract}

The applications of large language models (LLMs) in various biological domains have been explored recently, but their reasoning ability in complex biological systems, such as pathways, remains underexplored, which is crucial for predicting biological phenomena, formulating hypotheses, and designing experiments. This work explores the potential of LLMs in pathway reasoning. We introduce \benchname, a dataset with 5.1K complex pathway problems derived from real research, covering various biological contexts including natural dynamic changes, disturbances, additional intervention conditions, and multi-scale research targets. Our evaluation of methods such as CoT and graph-augmented reasoning, shows that LLMs struggle with pathway reasoning, especially in perturbed systems. To address this, we propose \modelname, an LLM agent that enhances reasoning through interactive subgraph-based navigation, enabling a more effective approach to handling the complexities of biological systems in a scientifically aligned manner. The dataset and code are available at \url{https://github.com/zhao-ht/BioMaze}.

\end{abstract}

\section{Introduction}
Large Language Models (LLMs) have recently demonstrated remarkable performance across scientific domains, including mathematics~\citep{yu2023metamath}, chemistry~\citep{liu2023multi, zhu2022torchdrug}, biology~\citep{hayes2024simulating, madani2020progen, ma2023retrieved}, and materials science~\citep{zheng2023shaping, park2024multi}. In biology, LLMs have shown promise in addressing complex tasks such as protein design~\citep{valentini2023promises, hosseini2024text2protein}, drug discovery~\citep{m2024augmenting, liu2023chatgpt}, clinical trial analysis~\citep{singhal2023large, jin2023matching}, and experiment design~\citep{ai4science2023impact, roohani2024biodiscoveryagent}.

\begin{figure*}[!t]
    \centering    
    \renewcommand{\thesubfigure}{} 
    \subfigure[]{\includegraphics[width=0.9\linewidth]{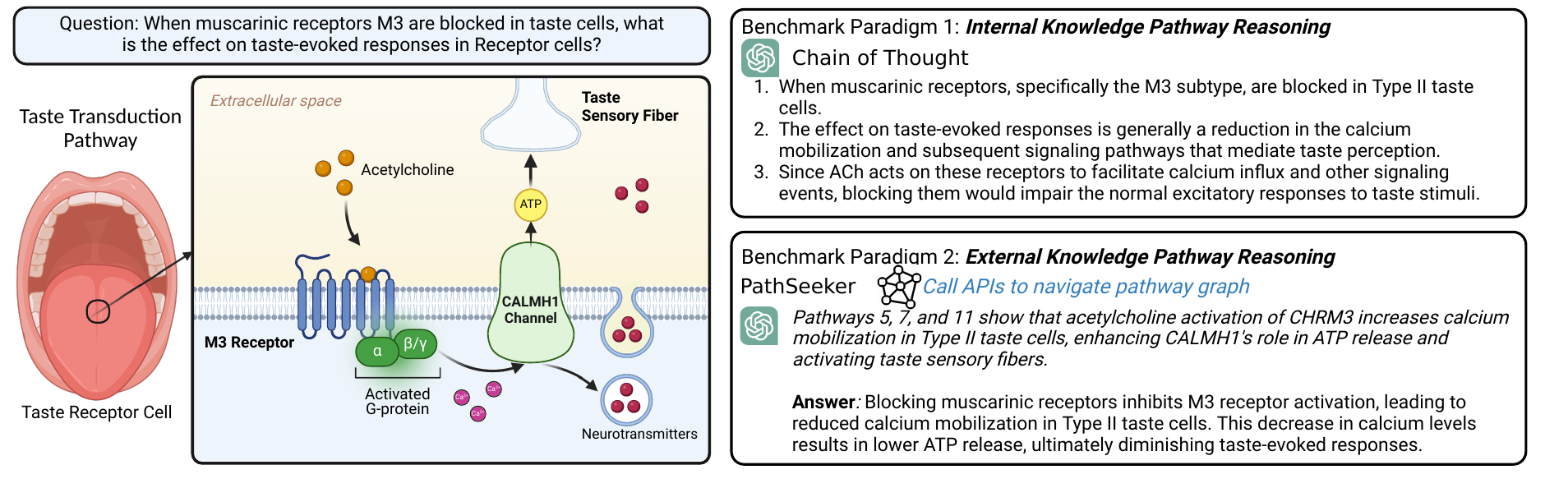}}\\
    \vspace{-2mm}
    \caption{Illustration of \benchname task and reasoning method with or without additional biological pathway graph data guidance. The task of \benchname focuses on reasoning about the effects and mechanisms of natural components or synthetic interventions on various downstream targets under different conditions through complex intermediate pathway processes.}
    \label{mainfig}
    \vspace{-2mm} 
\end{figure*}

Biological systems are composed of complex networks called pathways, involving genes, enzymes, substrates, and signaling molecules. Intervention in a single component—such as mutations or infections—can trigger multi-step cascades affecting other components within the organism. Despite the complex mechanisms in biological systems, deductive reasoning about the events within biological pathways can be carried out based on an understanding of the structure and function of these pathways. For example, based on pathway reasoning, it can be predicted that blocking muscarinic M3 receptors in taste cells will weaken taste responses in sensory fibers (see Figure \ref{mainfig}).

Pathway reasoning is essential for biologists to explain phenomena, form hypotheses, design experiments, and interpret results, and is a fundamental task in biology with broad applications across multiple disciplines, including systems biology, pharmacology, toxicology, cell biology, pathology, immunology, and biomedical engineering. The reasoning case in Figure \ref{mainfig} aid in toxicity analysis, experimental design, and taste disorders treatments.


Although LLMs have been explored in various biological applications, little research has focused on how LLMs can understand and reason through the intricate, multi-step processes inherent to complex biological systems. Considering the fundamental role of biological pathway reasoning, the potential applications of LLMs in many areas of biology would be questionable if LLMs cannot effectively perform this task.

In this study, we evaluate the reasoning abilities of LLMs in biological tasks through the lens of biological pathways. We explore their capabilities in key pathway reasoning paradigms, including functional understanding, dynamic changes, regulation, and intervention. To support these studies, we introduce a pathway benchmark, \benchname, which contains 5.1K high-quality, complex biological pathway problems derived from real research literature, such as PubMed~\citep{lu2011pubmed}. These problems are carefully curated and reviewed by human experts, covering the context of biological pathway research, including natural dynamic changes, disturbances and interventions, additional intervention conditions, as well as multi-scale research targets, such as single factors, interaction processes, and macro-level functions.

Based on \benchname, we compare various methods using LLMs for pathway reasoning, including chain-of-thought (CoT) and graph-augmented reasoning approaches ~\citep{li2023chain, sun2023think, he2024g}. The results demonstrate that LLMs struggle with pathway reasoning, particularly in causal inference scenarios where interventions and perturbations are introduced into the system. This challenge persists across all LLMs, from LLaMA 8B to GPT-4. The pathway knowledge of LLMs comes from the biological pathway corpus during pre-training, which lacks a structured organization, making it difficult to plan reasoning paths based on pathway structures. This leads to errors such as faulty steps and omissions, and difficulty in simulating the events in the pathway during interventions. Additionally, current graph-augmented reasoning methods lack the ability to efficiently leverage pathway graphs during reasoning.

To address these challenges, we propose a novel approach called \modelname, an LLM agent designed to emulate the way scientists reason using biological pathways. During inference, it interactively explores biological pathways by efficiently navigating global-local subgraphs based on demand. This interactive process establishes a mutually reinforcing relationship between inference and pathway browsing, helping LLMs utilize pathway data for reasoning. It addresses challenges such as interventions and perturbations, long reasoning chains, and errors.

\section{Related Work}

\textbf{Biological Scientific Question Answering} Previous studies have explored the potential of language models in the biological scientific domain \citep{lu2022learn, vilares2019head, jin2021disease, pal2022medmcqa}, such as biological scientific reading comprehension \citep{welbl2018constructing, jin2019pubmedqa} and question-answering \citep{krithara2023bioasq}. A few studies have examined language models' ability to complete biological pathways \citep{li2023chatpathway, park2023comparative, azam2024comprehensive}. Different from previous tasks, this work introduces biological pathway reasoning in realistic research scenarios. See Appendix \ref{appendix related work} for a more detailed comparison.

\textbf{Graph-augmented Language Model} Several studies have explored augmenting LLMs with graph data. In particular, some works enhance LLMs by encoding graph data as text \citep{ye2023natural, wang2024can, fatemi2023talk}, or tuning LLMs specifically for graph-based tasks \citep{liu2023one, tang2024graphgpt, he2024g, zhao2023gimlet, he2024unigraph}. 
Augmented LLMs have been applied to knowledge-based QA \citep{sun2023think, he2024g, li2023chain, jin2024graph, cheng2024call, edge2024local} and graph tasks like property prediction \citep{wang2024can, he2023harnessing}. Some studies leverage graph structures for complex reasoning tasks \citep{jiang2023structgpt, besta2024graph}. Most large graph databases use retrieval mechanisms \citep{he2024g, li2023chain}, while others employ LLMs as interactive agents for navigation \citep{sun2023think, jin2024graph, li2024graphreader}. This work introduces a more efficient agent-based approach using subgraph navigation and reasoning to improve pathway database exploration.

\section{Benchmark: \benchname}


\subsection{Dataset Creation}

\benchname is created by generating question-answer pairs from biological pathway research papers, which are then checked and filtered through a combination of automated methods and expert human review. The dataset creation process involves prompting large language models, with GPT-4 and LLaMA3.1-405B \citep{dubey2024llama} being selected for data generation in this study.

To gather relevant biological pathway questions in realistic scientific research contexts, particularly those involving interventions, the data for \benchname is sourced from over 6,000 biological pathway research papers. These studies include carefully designed experimental interventions supported by pathway mechanisms to observe biological system responses. After extracting detailed experimental observations and their contexts, we convert each one into either a True/False or open-ended question, depending on its content. Each question is paired with corresponding labeled answers.




We then apply multiple data filters and human expert reviews to ensure the accuracy and quality of the questions. The accuracy of each question is validated by comparing it with the content of the original paper. Question quality is ensured through several filters that remove questions that are poorly defined, ask for specific measurement values, query more than one fact, are trivial (with answers revealed in the question's context), or are unrelated to biological pathways.

Finally, all questions are reviewed by human experts based on quality dimensions and their judgment to ensure overall question quality. The passing rate for expert review is approximately 40\%. After applying all filters, \benchname contains 5.1k high-quality questions. More details are provided in Appendix \ref{appendix: data creation}.

The questions of \benchname cover a wide range of biological domains, as illustrated in Figure \ref{fig bio fan} (left).

\subsection{Reasoning Type Categories}

\begin{table*}[!t]
\tiny 
\scriptsize
\resizebox{\linewidth}{!}{
\begin{tabular}{ccm{5cm}c}
\toprule
Dimension &	Category &	Example (abbreviated) & Illustration\\
\midrule
\multirow{4}{*}{Inquiry Type}               & Normal     & What is the effect of \textcolor{blue}{AMPK activation} on SIRT1 activity in mouse skeletal muscle?  &   \begin{minipage}{0.2\textwidth}
                   \includegraphics[width=\linewidth]{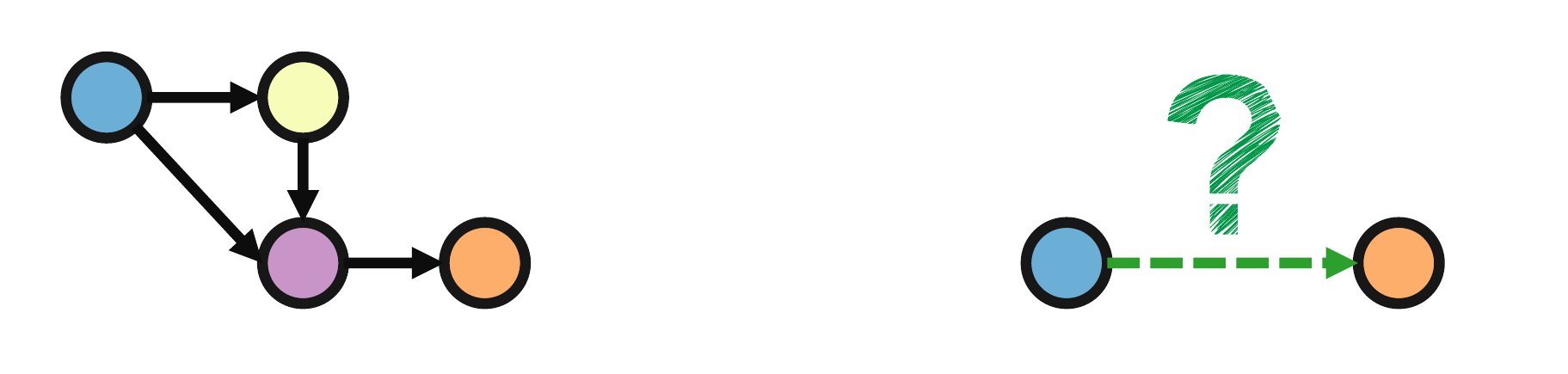} 
                \end{minipage}                           \\
                                      & Perturbed   & What is the effect of \textcolor{blue}{GogB-deficient Salmonella} on NFkappaB activation and proinflammatory responses in infected mice?    & \begin{minipage}{0.2\textwidth}
                   \includegraphics[width=\linewidth]{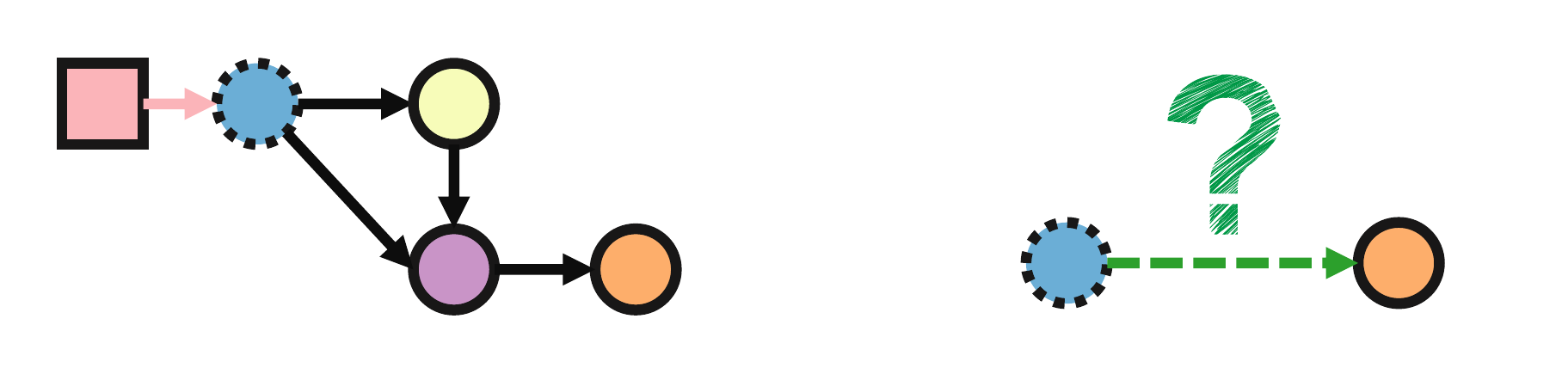} 
                \end{minipage}         \\

\midrule
\multirow{4}{*}{Extra Condition} & Natural       & How does apelin affect TNFalpha inhibition on brown adipogenesis?  & \begin{minipage}{0.2\textwidth}
                   \includegraphics[width=\linewidth]{fig/task_illustration_new_1.pdf} 
                \end{minipage}                                            \\
                                      & Intervened        & What is the role of BID in BAX activation in AIF-mediated necroptosis \textcolor{blue}{after MNNG treatment}?  & \begin{minipage}{0.2\textwidth}
                   \includegraphics[width=\linewidth]{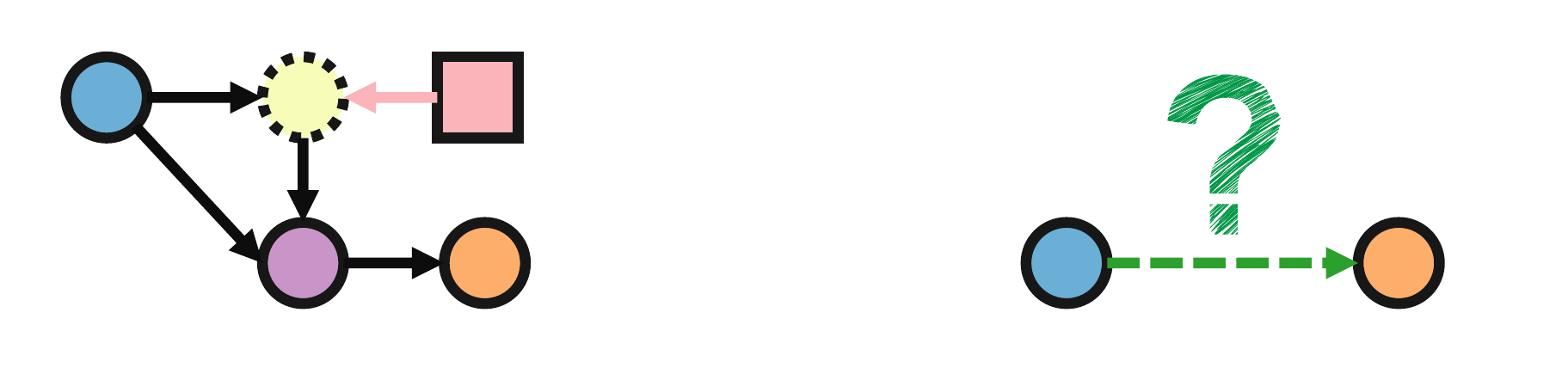} 
                \end{minipage}       \\ 

\midrule
\multirow{7}{*}{Investigation Target}               & Single       & What happens to \textcolor{blue}{AQP2} upon ADH stimulation? & \begin{minipage}{0.2\textwidth}
                   \includegraphics[width=\linewidth]{fig/task_illustration_new_1.pdf} 
                \end{minipage}                                                                    \\
                                      & Interaction & How does the influenza protein NS1 affect the \textcolor{blue}{activation of RIG-I by viral ssRNA}? & \begin{minipage}{0.2\textwidth}
                   \includegraphics[width=\linewidth]{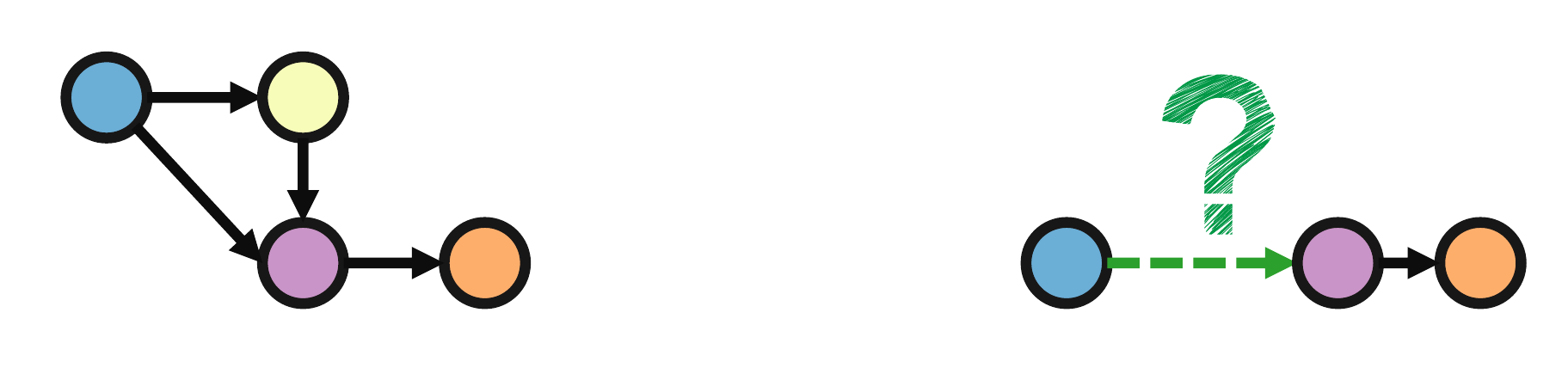} 
                \end{minipage}                             \\
                                      & Function    & What is the effect of losing 11beta-HSD2 from the fetus and fetally derived tissues on \textcolor{blue}{cerebellum development}? & \begin{minipage}{0.2\textwidth}
                   \includegraphics[width=\linewidth]{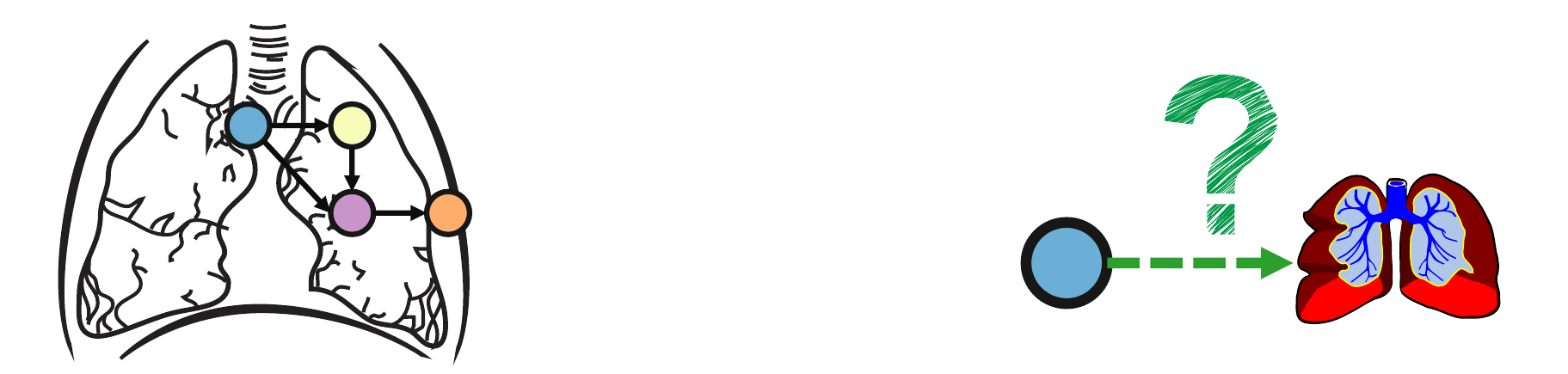} 
                \end{minipage} \\

\bottomrule
\end{tabular}}
\caption{Task example and causal illustration for each category.}
\label{table category example}
\vspace{-2mm}
\end{table*}



To study various research scenarios in biological pathways, such as natural dynamic changes, disturbances and interventions, and additional intervention conditions, as well as a multi-scale understanding of single factors, action processes, and macroscopic functions, we classify BioMaze tasks from three dimensions, namely inquiry type, extra condition, and investigation target, as shown in Table \ref{table category example}. More full question cases are in Appendix \ref{appendix data cases}. The distribution of the three dimensions' questions is shown in Figure \ref{fig bio fan} (right). We introduce each category of the dimensions below:

\begin{figure*}[!t]
    \centering    
    \renewcommand{\thesubfigure}{} 
    \subfigure[]{\includegraphics[width=0.36\linewidth]{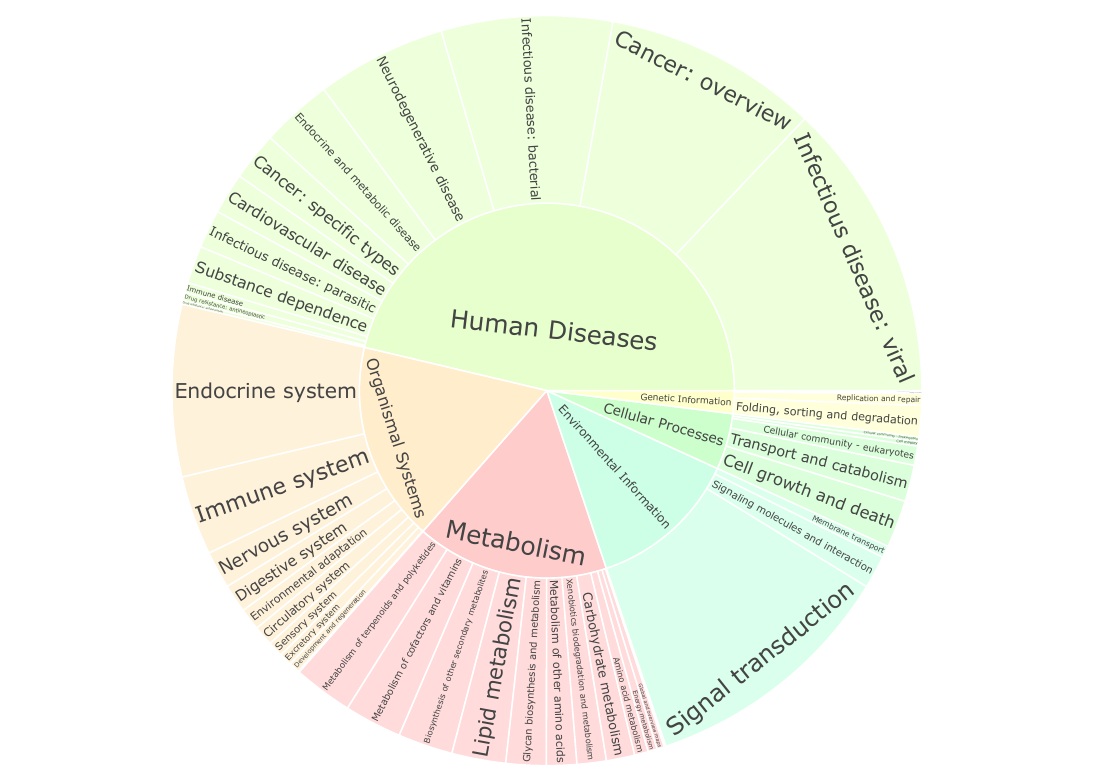}} 
    \subfigure[]{\includegraphics[width=0.36\linewidth]{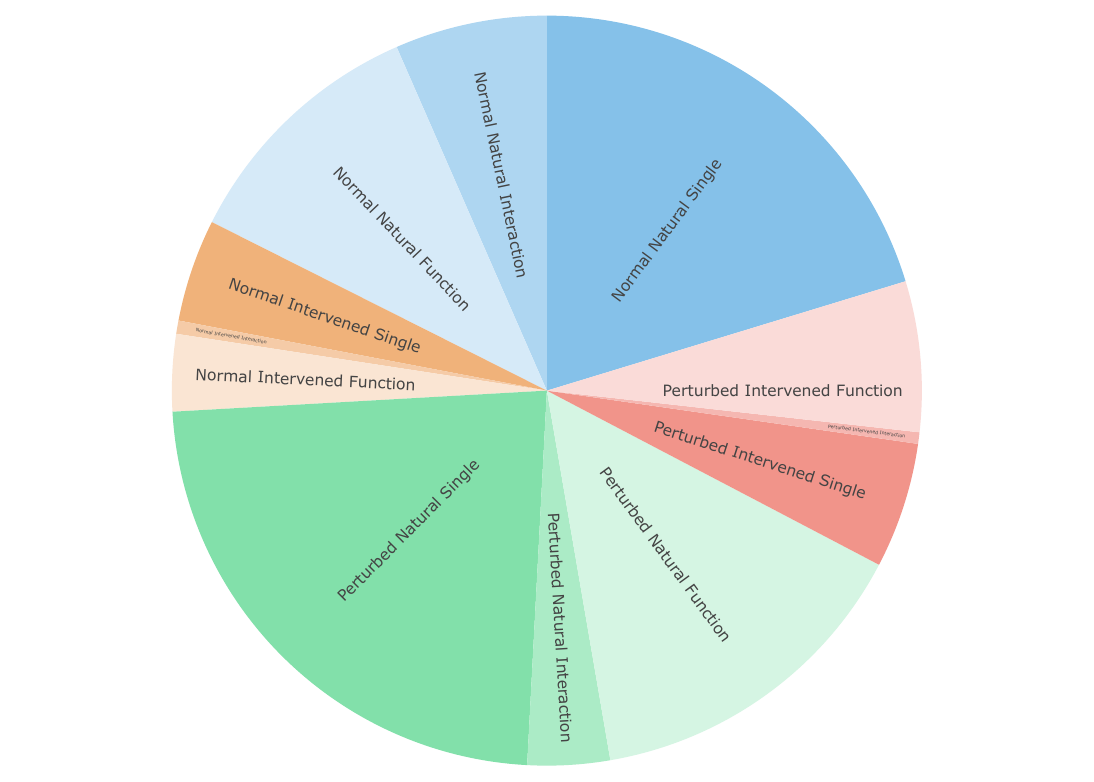}} 
    \vspace{-2mm}
    \caption{Dataset biological domain and reasoning type distribution. Left: \benchname covers six main domains: metabolism, genetic information processing, environmental information processing, cellular processes, organismal systems, and human diseases. Right: \benchname is categorized along three dimensions of reasoning types: inquiry type, extra condition, and investigation target.}
    \label{fig bio fan}
    \vspace{-2mm}
\end{figure*}



\textbf{Dimension 1: Inquiry Type} is the independent variable studied, which can be either \textbf{Normal Source}, involving the prediction of the effects of natural components in their normal state within a biological pathway, or \textbf{Perturbed Source}, which deals with predicting the effects of external interventions or treatments—such as mutations, infections, or experimentally introduced elements—on downstream targets within pathways. Normal Source tasks focus on understanding the fundamental mechanisms and natural dynamics of pathways, while Perturbed Source tasks examine the phenomenon under perturbation.

%





\textbf{Dimension 2: Extra Condition} refers to additional settings besides the independent variable. This could be the \textbf{Natural Condition}, where no additional treatments are applied, and the pathway operates under the organism's natural conditions, or the \textbf{Intervened Condition}, which assesses the impact of the inquiry source when the pathway has already been influenced by other factors, such as mutations or interventions. The Intervened Condition challenges the model by requiring it to deduce the system's behavior under unnatural conditions, thus increasing the reasoning difficulty.






\textbf{Dimension 3: Investigation Target} refers to the dependent variable in the question, which could be \textbf{Single Component as Target}, focusing on the effect of the source on a specific component within the pathway; \textbf{Components Interaction as Target}, examining the effect of the source on interactions between components within the pathway; or \textbf{Function as Target}, evaluating the effect of the source on broader biological functions or macro-level phenomena. The multi-scale targets address the reasoning of single components, downstream processes, or organism-wide outcomes.

\subsection{Pathway Graph Augmentated Reasoning}

Text-only reasoning methods, such as Chain-of-Thought (CoT), generate reasoning based on the inherent knowledge of LLMs. However, biological pathways present unique challenges due to their graph-structured nature. Consequently, using graph-augmented LLM reasoning is a natural approach for BioMaze. Providing explicit pathway graphs can serve as structural blueprints, enhancing reasoning from both knowledge and planning perspectives. We formalize this problem as follows:
\begin{equation}
\begin{aligned}
a = G (\mathcal{E}, o),
\end{aligned}
\end{equation}
where $G$ represents the language model, $\mathcal{E}$ denotes the task instruction (including the question), $o$ refers to the observation from the augment pathway graph database, and $a$ is the model output which could be the reasoning and answer.



\textbf{Pathway Graph Database} To augment LLMs with reasoning in biological pathways, we created a pathway graph database based on KEGG \citep{kanehisa2000kegg}, a collection of pathway maps on metabolism and various cellular and organismal functions widely-used resource among biologists. {We compiled all available pathway networks and maps from KEGG and integrated all of them into a single pathway graph database}. The statistics for the pathways are in Appendix \ref{appendix pathway statistic}. Each entry in the dataset is provided with a detailed description and function corpus. The graph is structured in triples as [\text{Head IDs}, \text{Tail IDs}, \textcolor{gray}{(\text{Relation Type}, \text{Biological Process IDs})}].




\textbf{Pathway Graph Database API:} When the language model accesses the pathway database, it may need to retrieve relevant triples from the pathway graph using APIs like $\operatorname{Search\_Node}$, $\operatorname{Search\_Edge}$, $\operatorname{Search\_Triple}$, and $\operatorname{Search\_Subgraph}$ \citep{sun2023think,li2023chain}. Our pathway database supports these core retrieval APIs, which are based on detailed descriptions and functional corpora. These APIs are essential for enabling various graph-augmented reasoning methods in LLMs. When the pathway graph $S$ is to be input to the LLM, they are encoded into text $o$ by $o=\operatorname{TripleToText}(\operatorname{DFSOrder}(S))$. The details of the API implementations are described in Appendix \ref{appendix subgraph retriever}.

\section{Method: Pathway Reasoning Agent \modelname}

\begin{figure*}[!t]
    \centering    
    \renewcommand{\thesubfigure}{} 
    \subfigure[]{\includegraphics[width=0.9\linewidth]{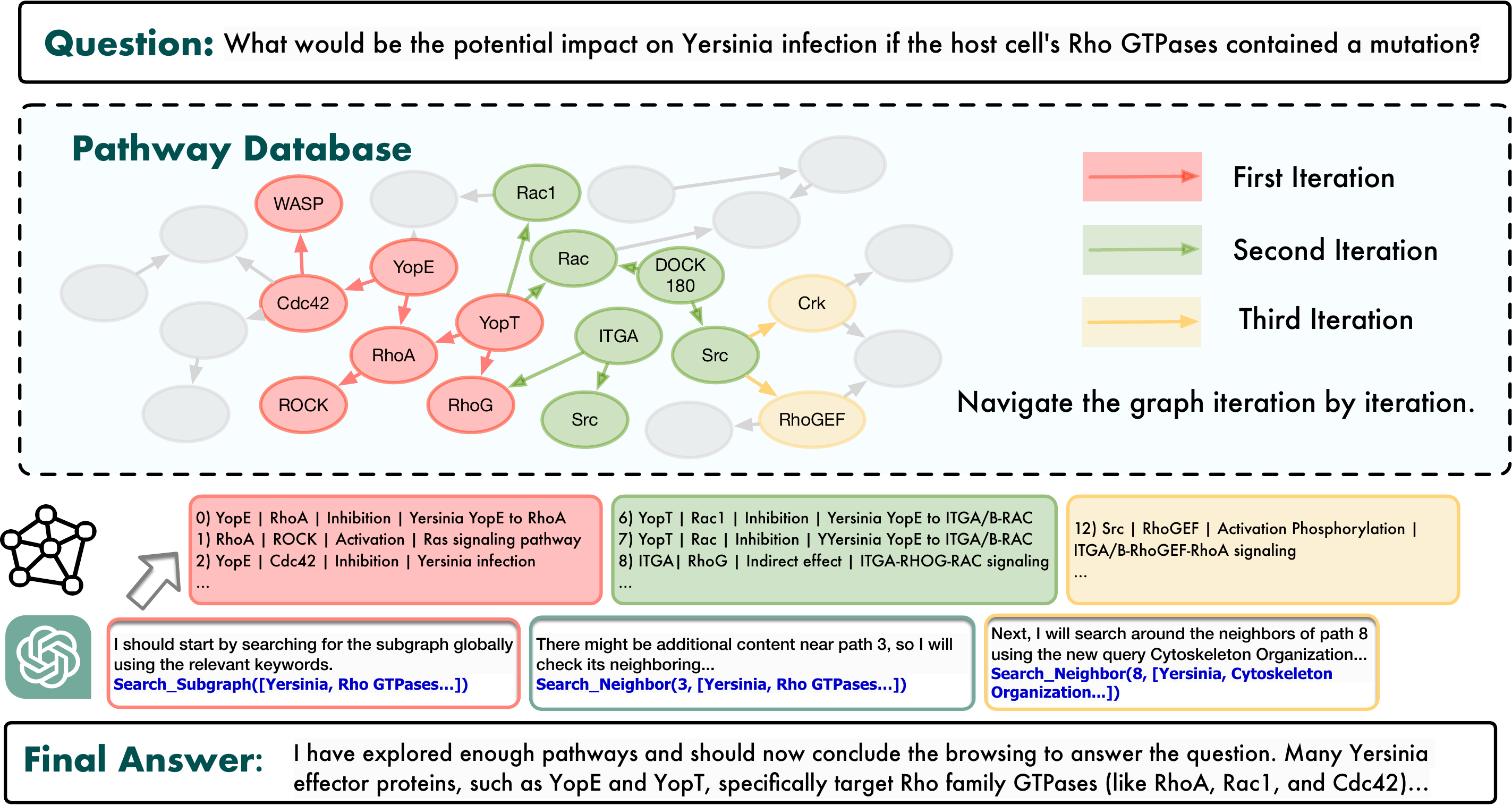}}
    \vspace{-2mm}
    \caption{\modelname allows interactive browsing of the pathway graph database by navigating through subgraphs. At each step, \modelname can perform either a global subgraph search or a local search around a previously explored pathway step. This functionality enables \modelname to fully leverage the augmented pathway graph database during biological pathway reasoning.}
    \label{fig agent method}
    \vspace{-2mm}
\end{figure*}

As we evaluated several graph-augmented reasoning methods, we found that current graph-augmentation methods' performance is limited by their ineffective utilization of the pathway graph database for reasoning. Inspired by how scientists browse pathway networks during reasoning, we propose \modelname, a reasoning agent method that can interactively conduct reasoning and take actions to perceive and navigate pathways using a web-like engine, along with flexible reasoning in each step, as shown in Figure \ref{fig agent method}.





At each step, $t$, the language agent $G$ can conduct reasoning by natural language thought, and takes an action step $a_t$, based on problem $\mathcal{E}$ (problem instructions) and previous observation-action trajectory $h_t=[o_1,a_1, \dots, o_{t-1}, a_{t-1}, o_t]$, 
\begin{equation}
  a_t =G(\mathcal{E}, h_t)  
\end{equation}

\textbf{Global and Local Subgraph Navigation} In addition to the global subgraph retriever $\operatorname{Search\_Subgraph}$, \modelname has access to an additional neighbor subgraph retriever, $\operatorname{Neighbor\_Subgraph}(\texttt{line\_id}, \texttt{query}, N)$, which retrieves an optimal connected subgraph of target size from the multi-hop neighbors of a previously observed pathway step $\texttt{line\_id}$.
\begin{equation}
\small
\begin{aligned}
\operatorname{Neighbor\_Subgraph}&(\texttt{line\_id}, \texttt{query}, N)= \\\underset{\substack{S \subseteq P_{id}, S \text { is connected }, |S|=N}}{\operatorname{argmax}} &\sum_{i \in V_S \cup E_S} \operatorname{score}(i, \texttt{query})
\end{aligned}
\end{equation}
Here, $P_{id}$ represents the multi-hop neighbors of the triple with $\texttt{line\_id}$. This allows \modelname to navigate the pathway graph database by either performing a global search or by exploring the multi-hop neighbors of an observed subgraph at each step. See Appendix for case \ref{appendix agent case}.

\textbf{Graph Encoding} In step $t$, the action taken by LLM agent get subgraph $S_t$ from environment, and the subgraph is encoded into text observation $o_t$ as following:
\begin{equation}
\small
\begin{aligned}
\hat{S}_t&=\operatorname{DFSOrder}\left(\operatorname{RemoveSeen}(S_t,[S_1, \dots, S_{t-1}])\right) \\
o_t&=\operatorname{TripleToOrderedText}\left(\hat{S}_t, \operatorname{TotalNum}\left([S_1, \dots, S_{t-1}]\right)\right)
\end{aligned}
\end{equation}
Function $\operatorname{RemoveSeen}$ eliminates triples from the $t$-th turn's subgraph that have been observed in previous turns, ensuring that each triple appears in the LLM's observations only once when first retrieved. This approach enhances content length efficiency and encourages the LLM to understand the whole navigation history rather than focusing solely on the most recent turn.

The function $\operatorname{TripleToOrderedText}$ convert ordered subgraph $\hat{S}_t$ into text in the following format: 
$ \text{Line ID)}  \text{ Head } |  \text{ Tail } | \text{ Relation and Biological Process}$. 
These global line IDs indicate the order of each triple across all turns, providing a unique reference for the LLM agent during local searches or reasoning. For the $t$-th turn's subgraph $S_t$, the ID starts at the total number of unique triples seen in previous history, given by $\operatorname{TotalNum}([S_1, \dots, S_{t-1}])$.

\textbf{Final Reasoning} As graph data browsing finishes, the final reasoning is conducted based on all the navigation history:
$$a_r =G(\mathcal{E}_r, [o_1, \dots, o_{T}])$$

\textbf{Graph Navigation Capacity}
The combination of global and local subgraph retrieval APIs empowers LLM agents to explore the entire network flexibly and efficiently. It allows the LLM to guide its exploration by adjusting both keywords and the root of the local subgraph, depending on the intermediate reasoning, offering stronger expressiveness than navigation methods like BFS, DFS, and various retrieval methods.


\section{Experiment}

\subsection{Baseline and Metric}

We evaluate the reasoning performance of LLMs on \benchname in both the unaugmented step-by-step reasoning and the pathway graph-augmented methods. We adopt reasoning method without graph augmentation Chain-of-Thought (CoT) \citep{wei2022chain,kojima2022large}, and methods with pathway graph augmentation: {Chain-of-Knowledge (CoK)} \citep{li2023chain}, {Think-of-Graph (ToG)} \citep{sun2023think}, and {G-Retriever} \citep{he2024g}. Details of baselines are in Appendix \ref{appendix baseline}.

For True/False tasks, we compute accuracy averaged across the True and False labels to account for label imbalance in the dataset. For open-ended tasks, the LLM is used to evaluate the accuracy of generated answers by comparing them to the ground truth and determining whether they are correct or incorrect. In this study, we use the LLaMA3.1-405B model as the evaluator, with five in-context examples. The performance of the evaluator is further analyzed in Appendix \ref{subsection evaluator quality}.

\subsection{Main Result}

\begin{table*}[!h]
\centering
\resizebox{1.0\linewidth}{!}{
\begin{tabular}{lc|cc|cc|ccc}
\toprule
  \multirow{2}{*}{\textbf{Method}} & \multirow{2}{*}{\textbf{w. Pathway Graph}} & \multicolumn{2}{c|}{\textbf{Inquiry Type}} &  \multicolumn{2}{c|}{\textbf{Extra Condition}} & \multicolumn{3}{c}{\textbf{Investigation Target}}\\

 &    & Normal & Perturbed & Natural & Intervened  & Single & Interaction & Function               \\
\midrule

\textcolor{brown}{\underline{\emph{GPT-3.5 }} }          &      &                           &                           &                           &                           &                           &                           &                           \\


Vanilla (0 Shot) & \multirow{4}{*}{\ding{55}}  & 57.92 & 54.60 & 56.99 & 54.88 & 59.91 & 55.63 & 56.68 \\
Vanilla (2 Shot) &  & 60.73 & 55.59 & 57.40 & 59.39 & 60.30 & 46.43 & 58.26 \\
CoT (0 Shot)     &  & 59.92 & \underline{61.48} & \textbf{62.74} & 51.00 & 57.69 & 56.75 & \underline{66.25} \\
CoT (2 Shot)     &  & \textbf{64.92} & 56.39 & 61.46 & 57.12 & 60.86 & \underline{61.01} & 59.92 \\
\midrule
ToG              & \multirow{4}{*}{\ding{51}} & 59.60 & 50.83 & 53.92 & \underline{62.50} & 53.40 & 60.00 & 55.21 \\
CoK              &  & 60.70 & 54.07 & 57.29 & 56.49 & 60.19 & 50.00 & 58.04 \\
G-Retriever      &  & \underline{64.14} & 59.32 & \underline{61.55} & 61.88 & \underline{61.53} & 59.00 & 62.60 \\
\modelname      &  & 63.55 & \textbf{63.93} & 57.48 & \textbf{62.74} & \textbf{62.85} & \textbf{64.73} & \textbf{68.13} \\

\midrule
\midrule
\textcolor{brown}{\underline{\emph{LLaMA3.1 8B }} }                   &                           &                           &                           &                           &                           &                           &                           \\

Vanilla (0 Shot) & \multirow{4}{*}{\ding{55}}  & 55.82 & 56.64 & 57.21 & 53.37 & 57.87 & 58.31 & 55.66 \\
Vanilla (2 Shot) &  & 55.92 & 58.88 & 60.04 & \underline{59.20} & \textbf{61.88} & 60.75 & 54.14 \\
CoT (0 Shot)     &  & 63.01 & 54.35 & 59.50 & 53.90 & 59.22 & 62.27 & 55.68 \\
CoT (2 Shot)     &  & 62.47 & \underline{57.73} & \underline{60.45} & 58.15 & 60.28 & 59.97 & \underline{59.47} \\
\midrule
ToG              & \multirow{4}{*}{\ding{51}}  & 58.99 & 55.31 & 56.67 & 58.35 & 56.79 & 57.85 & 57.10 \\
CoK              &  & 62.01 & 52.89 & 59.41 & 50.23 & 57.46 & \underline{62.57} & 55.43 \\
G-Retriever      &  & \underline{63.43} & 57.71 & 56.88 & 53.90 & \underline{61.54} & 60.01 & 59.10 \\
\modelname       &  & \textbf{63.69} & \textbf{60.25} & \textbf{62.30} & \textbf{62.91} & 61.27 & \textbf{63.19} & \textbf{63.99} \\
\bottomrule
\end{tabular}
}
\caption{Accuracy (\%) on \benchname True/False tasks (50\% corresponds to the random guessing baseline). The optimal results are in bold and the suboptimal ones are underlined.}
\label{table judge performance}
\vspace{-2mm}
\end{table*}

\begin{table*}[!h]
\centering
\resizebox{1.0\linewidth}{!}{
\begin{tabular}{lc|cc|cc|ccc}
\toprule
\multirow{2}{*}{\textbf{Method}} & \multirow{2}{*}{\textbf{w. Pathway Graph}} & \multicolumn{2}{c|}{\textbf{Inquiry Type}} &  \multicolumn{2}{c|}{\textbf{Extra Condition}} & \multicolumn{3}{c}{\textbf{Investigation Target}}\\

 &      & Normal & Perturbed & Natural & Intervened  & Single & Interaction & Function               \\
\midrule
\textcolor{brown}{\underline{\emph{GPT-3.5 }} }                 &                           &                           &                           &                           &                           &                           &                           \\


CoT (0 Shot) & \multirow{2}{*}{\ding{55}}  & 65.96 & 61.49 & \textbf{67.15} & 43.24 & 61.57 & 61.90 & \textbf{66.67} \\
CoT (2 Shot) &  & 65.43 & 59.08 & 63.29 & \underline{56.25} & \underline{67.76} & 66.29 & 53.26 \\
\midrule
ToG & \multirow{4}{*}{\ding{51}} & 65.52 & 59.86 & 64.71 & 48.65 & 66.39 & 64.00 & 56.17 \\
CoK &  & \textbf{70.27} & 54.00 & 63.87 & 50.00 & 62.93 & \underline{67.00} & 58.18 \\
G-Retriever &  & 65.19 & \underline{61.54} & 64.18 & 53.70 & \textbf{68.72} & 66.01 & 55.11 \\
\modelname  &  & \underline{67.51} & \textbf{64.33} & \underline{66.87} & \textbf{57.59} & 65.16 & \textbf{67.76} & \underline{65.79} \\

\midrule
\midrule

\textcolor{brown}{\underline{\emph{LLaMA3.1 8B }} }                    &                           &                           &                           &                           &                           &                           &                           \\


CoT (0 Shot) & \multirow{2}{*}{\ding{55}}  & \textbf{62.79} & \underline{57.19} & \textbf{61.50} & \underline{51.16} & \underline{62.77} & 62.50 & \textbf{55.31} \\
CoT (2 Shot) &  & 58.09 & 49.52 & 55.31 & 45.11 & 60.06 & 60.32 & 42.61 \\
\midrule
ToG &  \multirow{4}{*}{\ding{51}} & 52.14 & 49.48 & 52.05 & 43.60 & 53.33 & 55.24 & 45.68 \\
CoK &  & 60.55 & 55.23 & 59.12 & 50.63 & 61.12 & \underline{62.50} & 52.15 \\
G-Retriever &  & 53.83 & 51.19 & 53.27 & 48.10 & 57.79 & 55.44 & 46.52 \\
\modelname  &  & \underline{61.65} & \textbf{60.78} & \underline{61.30} & \textbf{60.60} & \textbf{64.14} & \textbf{64.43} & \underline{55.07} \\

\bottomrule
\end{tabular}
}
\caption{Accuracy (\%, evaluated by LLM) on \benchname open-ended tasks.  The optimal results are in bold and the suboptimal ones are underlined.}
\label{table generative performance}
\vspace{-2mm}
\end{table*}

We evaluate \modelname and baseline methods on \benchname, presenting results in Tables \ref{table judge performance} and \ref{table generative performance}. The comparison covers task dimensions including signal source, additional conditions, and target. The results lead to the following conclusions:

\textbf{LLMs struggle with biological pathway reasoning.} Pathway reasoning tasks in \benchname are tough for LLMs, with True/False accuracy slightly above random. Both CoT and graph-augmented reasoning reveal the gap between LLM capabilities and the complexity of biological systems.

\textbf{Perturbation query in \benchname presents significant challenges.} LLMs perform worse with perturbed inquiry settings than normal ones, especially in True/False and open-ended formats. This suggests that reasoning about biological pathways is harder in intervention scenarios, where the events are less aligned with common biological knowledge. answerable using established biological knowledge about how typical pathways work.

\textbf{Intervened conditions present greater reasoning challenges.} Like perturbations, intervention condition complicate reasoning by disrupting the biological system. These scenarios require more deductive reasoning, as they rely less on typical biological knowledge.

\textbf{Reasoning target brings diverse challenges for reasoning.} The Investigation target creates varied difficulties, causing inconsistent performance across models and reasoning methods. "Function as target" is the most difficult category.

\textbf{Pathway augmentation can enhance reasoning in biological systems, especially for intervention cases.} As shown in Tables \ref{table judge performance} and \ref{table generative performance}, reasoning methods with pathway augmentation, especially \modelname, outperform non-augmented approaches. \modelname consistently exceeds CoT across most question types and categories, regardless of the backbone model, highlighting the value of integrating biological pathways to enhance reasoning in biological systems. Additionally, \modelname outperforms other graph augmentation methods, demonstrating the effectiveness of its subgraph-based navigation approach. Notably, it reduces the performance gap between natural and intervened/perturbed groups, helping bridge the gap in pathway causal reasoning.

\subsection{Analysis}



\begin{figure*}[!t]
    \centering    
    \renewcommand{\thesubfigure}{} 
    \subfigure[]{\includegraphics[width=0.90\linewidth]{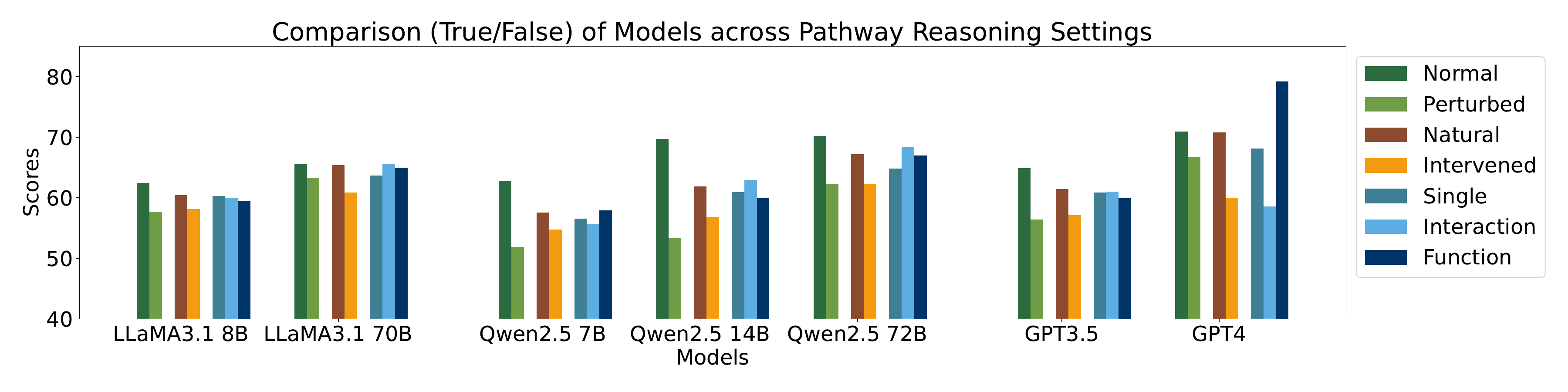}} \\
    \vspace{-10mm}
    \subfigure[]{\includegraphics[width=0.90\linewidth]{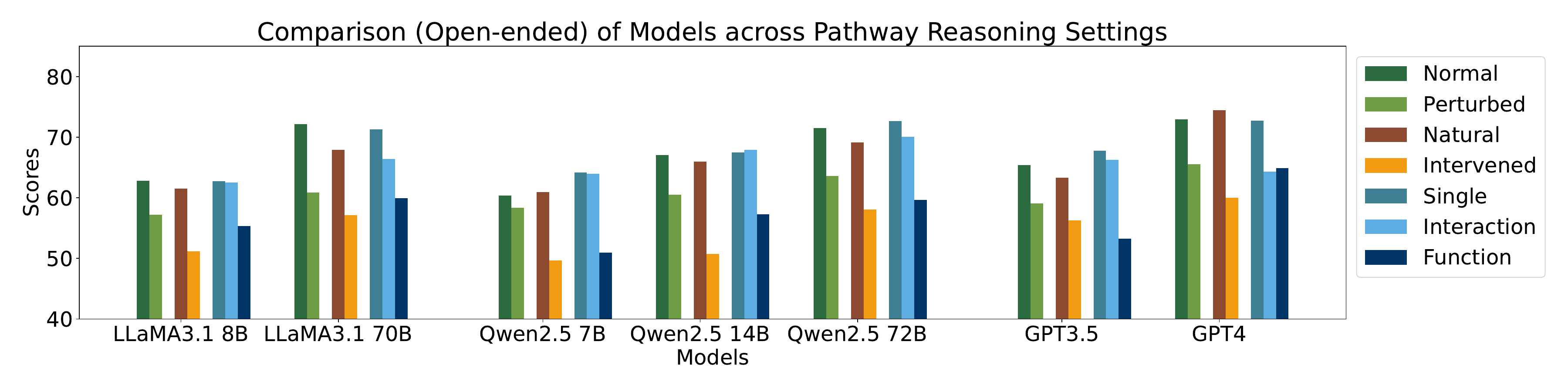}} 
    \vspace{-2mm}
    \caption{Comparison of the reasoning abilities of different LLMs for biological pathways. While overall performance improves with larger and more powerful models, a consistent gap exists between normal/perturbed and natural/intervened settings. This highlights the inherent limitations of LLMs in reasoning about the causal relationships within biological pathways.}
    \label{fig model comp}
    \vspace{-2mm}
\end{figure*}

\textbf{Backbone Ability for Pathway Reasoning} We compare the performance of different backbones in Figure \ref{fig model comp}. As the model size and capacity increase, overall performance improves, indicating a strong correlation between an LLM’s general reasoning ability and its performance in pathway reasoning. However, a consistent gap remains between the normal/perturbed and natural/intervened settings across different backbones. This underscores the inherent limitations of LLMs in reasoning about causal relationships within biological pathways.

\textbf{Reasoning Difficulty with Steps} To explore the relationship between task difficulty and reasoning steps in \benchname, we prompted LLaMA3.1-405B to explain its reasoning process based on the correct answer and pathway to get the step numbers.

\begin{figure}[!h]
    \centering    
    \renewcommand{\thesubfigure}{} 
    \subfigure[]
    {\includegraphics[width=0.45\linewidth]{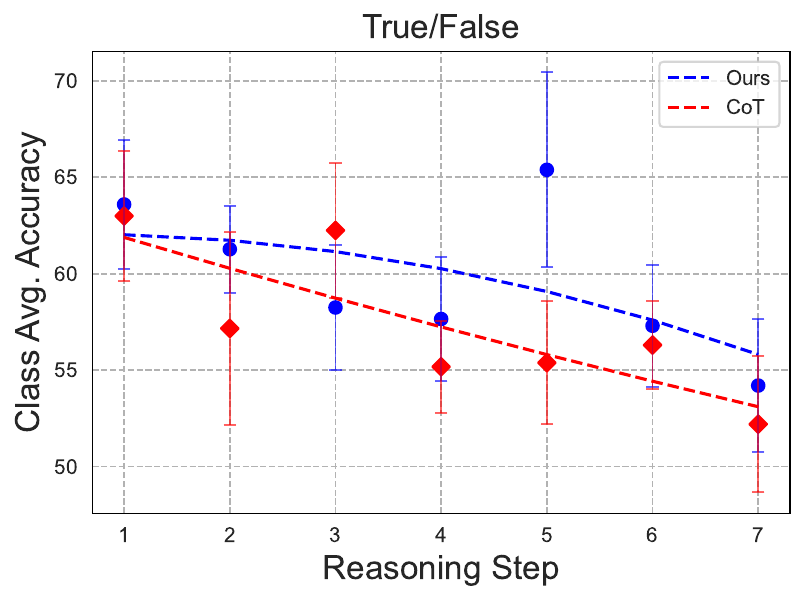}}{\includegraphics[width=0.45\linewidth]{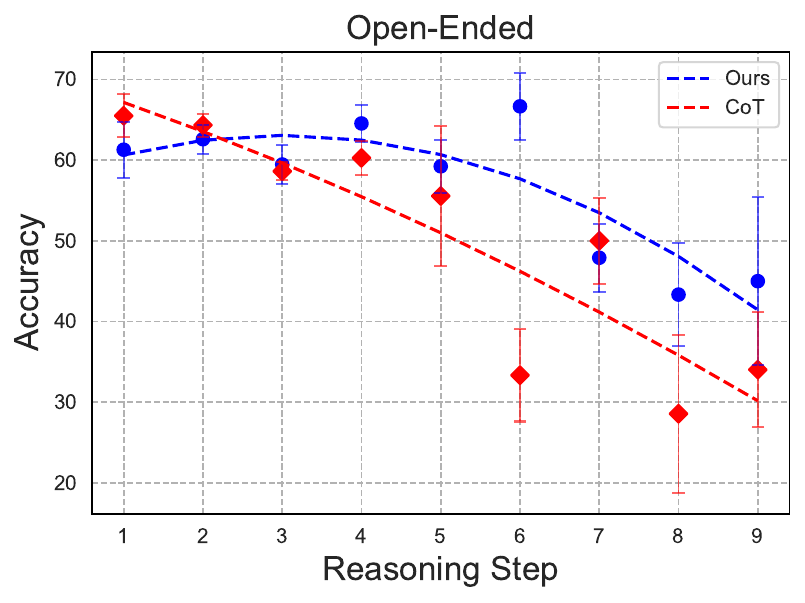}}

    \vspace{-2mm}
    \caption{Performance versus reasoning steps. LLMs face increasing difficulty in reasoning about biological systems as task complexity rises and requires more reasoning steps. In contrast, pathway augmentation significantly mitigates the drop of performance for tasks that involve more steps.}
    \label{fig reasoning step}
    \vspace{-2mm} 
\end{figure}

Figure \ref{fig reasoning step} shows that Chain-of-Thought (CoT) performance declines as reasoning steps increase, suggesting that more steps lead to higher reasoning difficulty. This supports our hypothesis that the complexity of the pathways is one factor of challenges in biological pathway reasoning.

Notably, \modelname's performance remains more consistent across different reasoning step counts. This suggests that augmenting LLMs with biological pathway information can mitigate the challenges of pathway reasoning, particularly when dealing with intricate intermediate processes.

\textbf{Failure Reasons Statistics} We analyze failed cases in biological pathway reasoning, covering CoT and \modelname, and classify the failures into: (1) \textbf{Unresolved Conclusion (UC)} For cases where the model fails to provide a definitive answer, indicating uncertainty or belief that the answer is unknown.
(2) \textbf{Incomplete Answer (IA)}  When the response lacks essential details, such as missing the requested effects or other key elements.
(3) \textbf{Omission in Reasoning (OR)}  For errors where critical pathway steps in the question's biological process are left out, causing the final answer to be incorrect.
(4) \textbf{Faulty in Reasoning (FR)} When the reasoning path is correct, but there are significant errors in deducing the events within that pathway. We manually classify 200 random samples from these error cases to approximate the overall error cases, with a professional biology Ph.D. student.

\begin{figure}[!t]
    \centering    
    \renewcommand{\thesubfigure}{} 
    \subfigure[]{\includegraphics[width=0.45\linewidth]{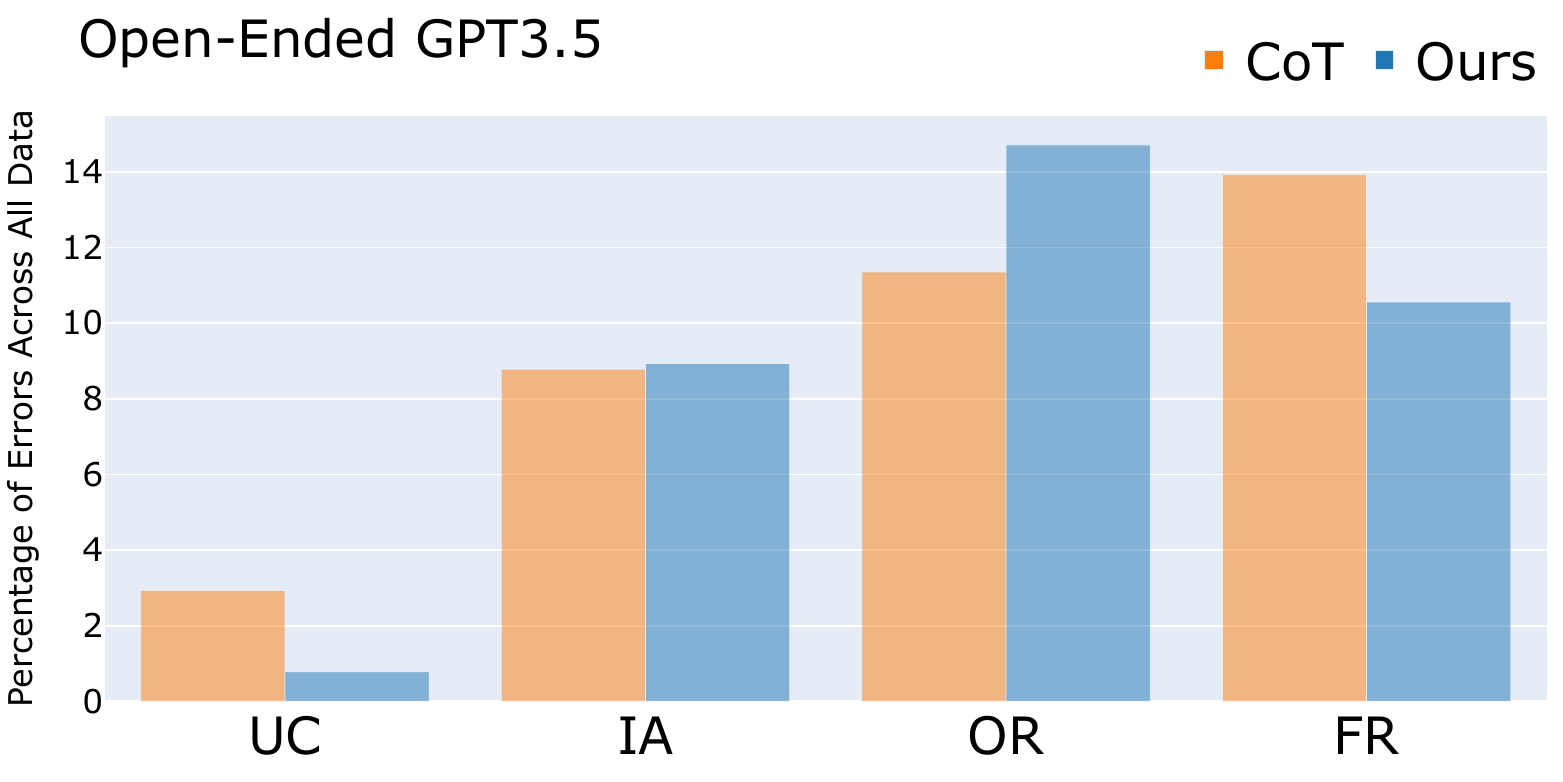}}
    {\includegraphics[width=0.45\linewidth]{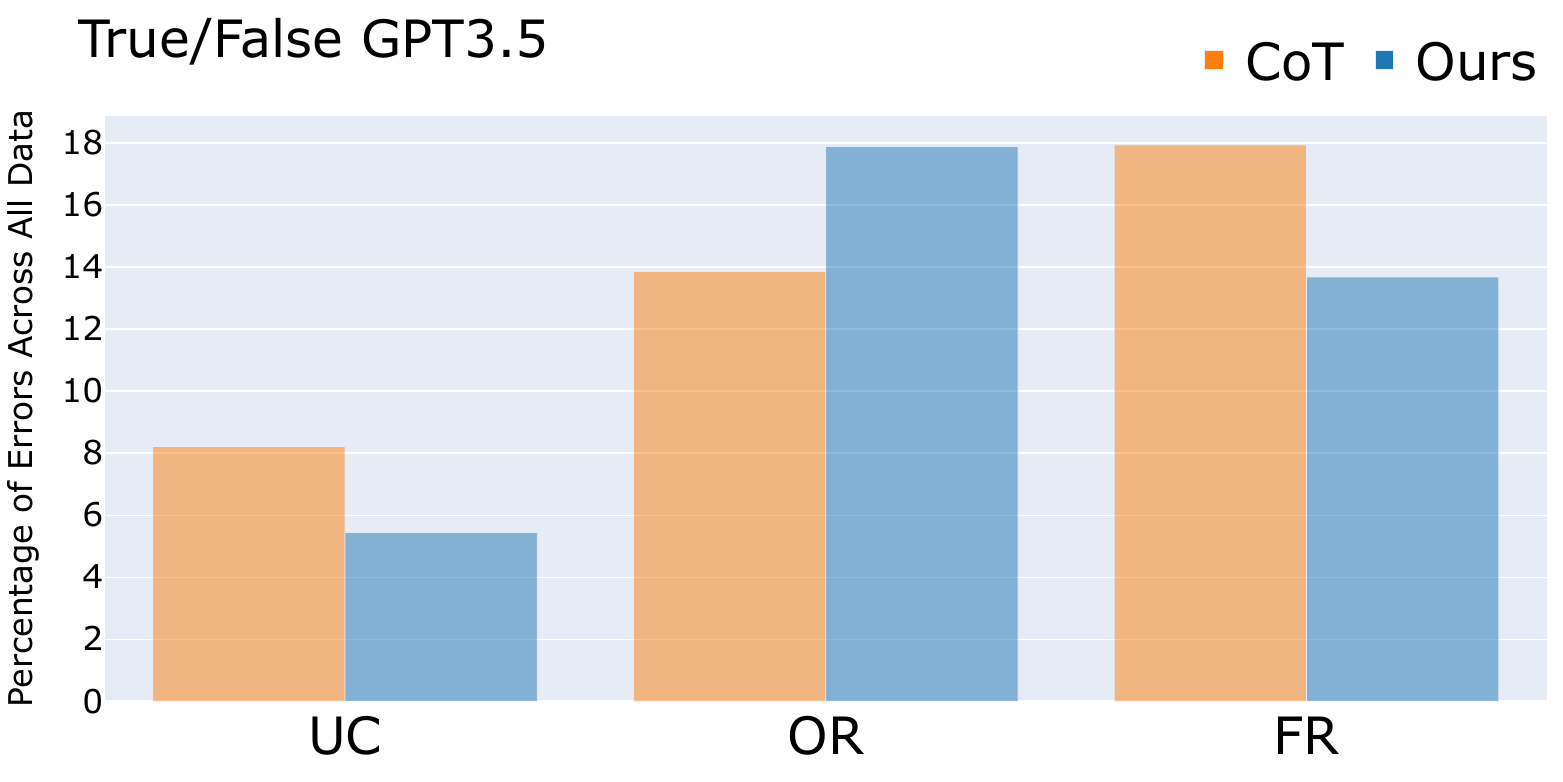}}
    \vspace{-2mm}
    \caption{Error analysis for CoT reasoning and reasoning with pathway augmentation (our method \modelname). The primary cause of errors in (CoT) reasoning for biological systems is due to both faulty reasoning and omissions in reasoning. When pathway augmentation is applied, omissions in reasoning become the predominant issue, but the rate of faulty reasoning is significantly reduced, thereby improving the overall reasoning accuracy of LLMs in biological systems.}
    \label{fig error analysis}
    \vspace{-2mm} 
\end{figure}

The results in Figure \ref{fig error analysis} show that in both True/False and open-ended tasks, the main error in CoT reasoning is faulty reasoning, where LLMs correctly identify the biological pathway but misinterpret the events within it. Another key error is omission, where critical steps or branches of the pathway are overlooked. This highlights the challenges LLMs face in reasoning about biological pathways, due to both knowledge gaps and difficulties in deductive reasoning.

\modelname reduces faulty reasoning by providing pathway graphs, improving accuracy. However, omissions remain a challenge, often due to limitations in the pathway database and browsing issues. With pathways available, LLMs are more confident and less fail in drawing conclusions.

\begin{minipage}{\textwidth}
\begin{minipage}[b]{0.48\textwidth}
\makeatletter\def\@captype{table}
\centering
\resizebox{0.84\linewidth}{!}
{
\begin{tabular}{lccccc}
\toprule
Agent \#Steps          & 1-4   & 4-6   & 6-8   & 8-10 & $\ge$10  \\
\midrule
True / False     & 0.91 &	50.14	& 26.58 &	12.66 &	9.70 \\
Open-Ended & 1.45 &	52.44 &	25.69 &	13.97 &	6.46\\
\bottomrule
\end{tabular}
}
\vspace{-2mm}
\caption{Agent steps distribution (\%) of \modelname during task completion.}
\label{table agent steps}
\vspace{-2mm}

\end{minipage}
\hspace{0.3cm}
\begin{minipage}[b]{0.48\textwidth}
\makeatletter\def\@captype{table}
\centering
\resizebox{0.5\linewidth}{!}{
\begin{tabular}{lcc}
\toprule
   Task Type       & Global & Local \\
\midrule
True / False     & 1.47 &	3.62  \\
Open-Ended & 1.57 &	3.43  \\
\bottomrule
\end{tabular}
}
\vspace{-2mm}
\caption{Average API usage times of \modelname during task completion.}
\label{table API usage}
\vspace{-2mm}
\end{minipage}
\end{minipage}

\subsection{Method Analysis and Ablation Study}


\textbf{API Usage and Step Distribution} We analyze \modelname's agent behavior by reviewing task steps and API usage. Tables \ref{table agent steps} and \ref{table API usage} show that most tasks are completed in six or fewer steps, though some require over ten due to missing pathway data. On average, the agent performs 1.5 global searches and over three local navigations per task, indicating frequent subgraph exploration.

\textbf{Ablation Study} To evaluate \modelname's components, we perform ablation studies, with results for LLaMA3-8B in Table \ref{table ablation}. The most important component is FinalReaser; without it, the agent's answers degrade due to long task history. The local search API is also crucial for efficient graph navigation, and the graph encoding method improves performance, highlighting the value of encoding graph data for sequential models.

\begin{table*}[!h]
\centering
\resizebox{0.98\linewidth}{!}{
\begin{tabular}{lcccccc}
\toprule
  Task Type        & \modelname & w.o. RemoveSeen & w.o. DFSOrder & w.o. TripleToText  & w.o. Local search & w.o. FinalReasoner \\
\midrule
True / False     & \textbf{61.87} &	57.48 &	58.60 &	58.32 &	57.78 &	56.97             \\
Open-Ended & \textbf{61.21} &	58.96 &	55.82 &	57.06 &	57.46 &	58.25         \\
\bottomrule
\end{tabular}
}
\caption{Ablation Study of \modelname.}
\label{table ablation}
\vspace{-2mm}
\end{table*}




\section{Conclusion}
In this study, we introduce \benchname, a benchmark designed to evaluate LLMs' ability to understand and reason about biological pathways. Extensive evaluations using \benchname, incorporating advanced methods like CoT and graph-augmented approaches, show that LLMs struggle with understanding pathway mechanisms. We also propose \modelname, a novel LLM agent that uses interactive subgraph exploration to enhance reasoning in biological pathways.

\bibliography{ref}

\begin{thebibliography}{52}
\providecommand{\natexlab}[1]{#1}
\providecommand{\url}[1]{\texttt{#1}}
\expandafter\ifx\csname urlstyle\endcsname\relax
  \providecommand{\doi}[1]{doi: #1}\else
  \providecommand{\doi}{doi: \begingroup \urlstyle{rm}\Url}\fi

\bibitem[AI4Science \& Quantum(2023)AI4Science and Quantum]{ai4science2023impact}
Microsoft~Research AI4Science and Microsoft~Azure Quantum.
\newblock The impact of large language models on scientific discovery: a preliminary study using gpt-4.
\newblock \emph{arXiv preprint arXiv:2311.07361}, 2023.

\bibitem[Azam et~al.(2024)Azam, Chen, Arowolo, Liu, Popescu, and Xu]{azam2024comprehensive}
Muhammad Azam, Yibo Chen, Micheal~Olaolu Arowolo, Haowang Liu, Mihail Popescu, and Dong Xu.
\newblock A comprehensive evaluation of large language models in mining gene relations and pathway knowledge.
\newblock \emph{Quantitative Biology}, 2024.

\bibitem[Besta et~al.(2024)Besta, Blach, Kubicek, Gerstenberger, Podstawski, Gianinazzi, Gajda, Lehmann, Niewiadomski, Nyczyk, et~al.]{besta2024graph}
Maciej Besta, Nils Blach, Ales Kubicek, Robert Gerstenberger, Michal Podstawski, Lukas Gianinazzi, Joanna Gajda, Tomasz Lehmann, Hubert Niewiadomski, Piotr Nyczyk, et~al.
\newblock Graph of thoughts: Solving elaborate problems with large language models.
\newblock In \emph{Proceedings of the AAAI Conference on Artificial Intelligence}, volume~38, pp.\  17682--17690, 2024.

\bibitem[Cheng et~al.(2024)Cheng, Zhuang, Xu, Yang, Zhang, Qin, Huang, Chen, Lin, Zhang, et~al.]{cheng2024call}
Sitao Cheng, Ziyuan Zhuang, Yong Xu, Fangkai Yang, Chaoyun Zhang, Xiaoting Qin, Xiang Huang, Ling Chen, Qingwei Lin, Dongmei Zhang, et~al.
\newblock Call me when necessary: Llms can efficiently and faithfully reason over structured environments.
\newblock \emph{arXiv preprint arXiv:2403.08593}, 2024.

\bibitem[Dubey et~al.(2024)Dubey, Jauhri, Pandey, Kadian, Al-Dahle, Letman, Mathur, Schelten, Yang, Fan, et~al.]{dubey2024llama}
Abhimanyu Dubey, Abhinav Jauhri, Abhinav Pandey, Abhishek Kadian, Ahmad Al-Dahle, Aiesha Letman, Akhil Mathur, Alan Schelten, Amy Yang, Angela Fan, et~al.
\newblock The llama 3 herd of models.
\newblock \emph{arXiv preprint arXiv:2407.21783}, 2024.

\bibitem[Edge et~al.(2024)Edge, Trinh, Cheng, Bradley, Chao, Mody, Truitt, and Larson]{edge2024local}
Darren Edge, Ha~Trinh, Newman Cheng, Joshua Bradley, Alex Chao, Apurva Mody, Steven Truitt, and Jonathan Larson.
\newblock From local to global: A graph rag approach to query-focused summarization.
\newblock \emph{arXiv preprint arXiv:2404.16130}, 2024.

\bibitem[Fatemi et~al.(2023)Fatemi, Halcrow, and Perozzi]{fatemi2023talk}
Bahare Fatemi, Jonathan Halcrow, and Bryan Perozzi.
\newblock Talk like a graph: Encoding graphs for large language models.
\newblock \emph{arXiv preprint arXiv:2310.04560}, 2023.

\bibitem[Hayes et~al.(2024)Hayes, Rao, Akin, Sofroniew, Oktay, Lin, Verkuil, Tran, Deaton, Wiggert, et~al.]{hayes2024simulating}
Tomas Hayes, Roshan Rao, Halil Akin, Nicholas~J Sofroniew, Deniz Oktay, Zeming Lin, Robert Verkuil, Vincent~Q Tran, Jonathan Deaton, Marius Wiggert, et~al.
\newblock Simulating 500 million years of evolution with a language model.
\newblock \emph{bioRxiv}, pp.\  2024--07, 2024.

\bibitem[He et~al.(2023)He, Bresson, Laurent, Perold, LeCun, and Hooi]{he2023harnessing}
Xiaoxin He, Xavier Bresson, Thomas Laurent, Adam Perold, Yann LeCun, and Bryan Hooi.
\newblock Harnessing explanations: Llm-to-lm interpreter for enhanced text-attributed graph representation learning.
\newblock \emph{arXiv preprint arXiv:2305.19523}, 2023.

\bibitem[He et~al.(2024)He, Tian, Sun, Chawla, Laurent, LeCun, Bresson, and Hooi]{he2024g}
Xiaoxin He, Yijun Tian, Yifei Sun, Nitesh~V Chawla, Thomas Laurent, Yann LeCun, Xavier Bresson, and Bryan Hooi.
\newblock G-retriever: Retrieval-augmented generation for textual graph understanding and question answering.
\newblock \emph{arXiv preprint arXiv:2402.07630}, 2024.

\bibitem[He \& Hooi(2024)He and Hooi]{he2024unigraph}
Yufei He and Bryan Hooi.
\newblock Unigraph: Learning a cross-domain graph foundation model from natural language.
\newblock \emph{arXiv preprint arXiv:2402.13630}, 2024.

\bibitem[Hegde et~al.(2015)Hegde, Indyk, and Schmidt]{hegde2015nearly}
Chinmay Hegde, Piotr Indyk, and Ludwig Schmidt.
\newblock A nearly-linear time framework for graph-structured sparsity.
\newblock In \emph{International Conference on Machine Learning}, pp.\  928--937. PMLR, 2015.

\bibitem[Hosseini et~al.(2024)Hosseini, Zhang, and Xie]{hosseini2024text2protein}
Ramtin Hosseini, Siyang Zhang, and Pengtao Xie.
\newblock Text2protein: A generative model for designated protein design on given description.
\newblock 2024.

\bibitem[Jiang et~al.(2023)Jiang, Zhou, Dong, Ye, Zhao, and Wen]{jiang2023structgpt}
Jinhao Jiang, Kun Zhou, Zican Dong, Keming Ye, Wayne~Xin Zhao, and Ji-Rong Wen.
\newblock Structgpt: A general framework for large language model to reason over structured data.
\newblock \emph{arXiv preprint arXiv:2305.09645}, 2023.

\bibitem[Jin et~al.(2024)Jin, Xie, Zhang, Roy, Zhang, Wang, Meng, and Han]{jin2024graph}
Bowen Jin, Chulin Xie, Jiawei Zhang, Kashob~Kumar Roy, Yu~Zhang, Suhang Wang, Yu~Meng, and Jiawei Han.
\newblock Graph chain-of-thought: Augmenting large language models by reasoning on graphs.
\newblock \emph{arXiv preprint arXiv:2404.07103}, 2024.

\bibitem[Jin et~al.(2021)Jin, Pan, Oufattole, Weng, Fang, and Szolovits]{jin2021disease}
Di~Jin, Eileen Pan, Nassim Oufattole, Wei-Hung Weng, Hanyi Fang, and Peter Szolovits.
\newblock What disease does this patient have? a large-scale open domain question answering dataset from medical exams.
\newblock \emph{Applied Sciences}, 11\penalty0 (14):\penalty0 6421, 2021.

\bibitem[Jin et~al.(2019)Jin, Dhingra, Liu, Cohen, and Lu]{jin2019pubmedqa}
Qiao Jin, Bhuwan Dhingra, Zhengping Liu, William~W Cohen, and Xinghua Lu.
\newblock Pubmedqa: A dataset for biomedical research question answering.
\newblock \emph{arXiv preprint arXiv:1909.06146}, 2019.

\bibitem[Jin et~al.(2023)Jin, Wang, Floudas, Chen, Gong, Bracken-Clarke, Xue, Yang, Sun, and Lu]{jin2023matching}
Qiao Jin, Zifeng Wang, Charalampos~S Floudas, Fangyuan Chen, Changlin Gong, Dara Bracken-Clarke, Elisabetta Xue, Yifan Yang, Jimeng Sun, and Zhiyong Lu.
\newblock Matching patients to clinical trials with large language models.
\newblock \emph{ArXiv}, 2023.

\bibitem[Kanehisa \& Goto(2000)Kanehisa and Goto]{kanehisa2000kegg}
Minoru Kanehisa and Susumu Goto.
\newblock Kegg: kyoto encyclopedia of genes and genomes.
\newblock \emph{Nucleic acids research}, 28\penalty0 (1):\penalty0 27--30, 2000.

\bibitem[Kojima et~al.(2022)Kojima, Gu, Reid, Matsuo, and Iwasawa]{kojima2022large}
Takeshi Kojima, Shixiang~Shane Gu, Machel Reid, Yutaka Matsuo, and Yusuke Iwasawa.
\newblock Large language models are zero-shot reasoners.
\newblock \emph{Advances in neural information processing systems}, 35:\penalty0 22199--22213, 2022.

\bibitem[Krithara et~al.(2023)Krithara, Nentidis, Bougiatiotis, and Paliouras]{krithara2023bioasq}
Anastasia Krithara, Anastasios Nentidis, Konstantinos Bougiatiotis, and Georgios Paliouras.
\newblock Bioasq-qa: A manually curated corpus for biomedical question answering.
\newblock \emph{Scientific Data}, 10\penalty0 (1):\penalty0 170, 2023.

\bibitem[Li et~al.(2024)Li, He, Guo, Bu, Bai, Liu, Liu, Qu, Li, Ouyang, et~al.]{li2024graphreader}
Shilong Li, Yancheng He, Hangyu Guo, Xingyuan Bu, Ge~Bai, Jie Liu, Jiaheng Liu, Xingwei Qu, Yangguang Li, Wanli Ouyang, et~al.
\newblock Graphreader: Building graph-based agent to enhance long-context abilities of large language models.
\newblock \emph{arXiv preprint arXiv:2406.14550}, 2024.

\bibitem[Li et~al.(2023{\natexlab{a}})Li, Zhao, Chia, Ding, Joty, Poria, and Bing]{li2023chain}
Xingxuan Li, Ruochen Zhao, Yew~Ken Chia, Bosheng Ding, Shafiq Joty, Soujanya Poria, and Lidong Bing.
\newblock Chain-of-knowledge: Grounding large language models via dynamic knowledge adapting over heterogeneous sources.
\newblock \emph{arXiv preprint arXiv:2305.13269}, 2023{\natexlab{a}}.

\bibitem[Li et~al.(2023{\natexlab{b}})Li, Xu, Zhao, Guo, and Liu]{li2023chatpathway}
Yanjing Li, Hannan Xu, Haiteng Zhao, Hongyu Guo, and Shengchao Liu.
\newblock Chatpathway: Conversational large language models for biology pathway detection.
\newblock In \emph{NeurIPS 2023 AI for Science Workshop}, 2023{\natexlab{b}}.

\bibitem[Liu et~al.(2023{\natexlab{a}})Liu, Feng, Kong, Liang, Tao, Chen, and Zhang]{liu2023one}
Hao Liu, Jiarui Feng, Lecheng Kong, Ningyue Liang, Dacheng Tao, Yixin Chen, and Muhan Zhang.
\newblock One for all: Towards training one graph model for all classification tasks.
\newblock \emph{arXiv preprint arXiv:2310.00149}, 2023{\natexlab{a}}.

\bibitem[Liu et~al.(2023{\natexlab{b}})Liu, Nie, Wang, Lu, Qiao, Liu, Tang, Xiao, and Anandkumar]{liu2023multi}
Shengchao Liu, Weili Nie, Chengpeng Wang, Jiarui Lu, Zhuoran Qiao, Ling Liu, Jian Tang, Chaowei Xiao, and Animashree Anandkumar.
\newblock Multi-modal molecule structure--text model for text-based retrieval and editing.
\newblock \emph{Nature Machine Intelligence}, 5\penalty0 (12):\penalty0 1447--1457, 2023{\natexlab{b}}.

\bibitem[Liu et~al.(2023{\natexlab{c}})Liu, Wang, Yang, Wang, Liu, Guo, and Xiao]{liu2023chatgpt}
Shengchao Liu, Jiongxiao Wang, Yijin Yang, Chengpeng Wang, Ling Liu, Hongyu Guo, and Chaowei Xiao.
\newblock Chatgpt-powered conversational drug editing using retrieval and domain feedback.
\newblock \emph{arXiv preprint arXiv:2305.18090}, 2023{\natexlab{c}}.

\bibitem[Lu et~al.(2022)Lu, Mishra, Xia, Qiu, Chang, Zhu, Tafjord, Clark, and Kalyan]{lu2022learn}
Pan Lu, Swaroop Mishra, Tanglin Xia, Liang Qiu, Kai-Wei Chang, Song-Chun Zhu, Oyvind Tafjord, Peter Clark, and Ashwin Kalyan.
\newblock Learn to explain: Multimodal reasoning via thought chains for science question answering.
\newblock \emph{Advances in Neural Information Processing Systems}, 35:\penalty0 2507--2521, 2022.

\bibitem[Lu(2011)]{lu2011pubmed}
Zhiyong Lu.
\newblock Pubmed and beyond: a survey of web tools for searching biomedical literature.
\newblock \emph{Database}, 2011:\penalty0 baq036, 2011.

\bibitem[M.~Bran et~al.(2024)M.~Bran, Cox, Schilter, Baldassari, White, and Schwaller]{m2024augmenting}
Andres M.~Bran, Sam Cox, Oliver Schilter, Carlo Baldassari, Andrew~D White, and Philippe Schwaller.
\newblock Augmenting large language models with chemistry tools.
\newblock \emph{Nature Machine Intelligence}, pp.\  1--11, 2024.

\bibitem[Ma et~al.(2023)Ma, Zhao, Zheng, Xin, Li, Wu, Deng, Lu, Liu, and Kong]{ma2023retrieved}
Chang Ma, Haiteng Zhao, Lin Zheng, Jiayi Xin, Qintong Li, Lijun Wu, Zhihong Deng, Yang Lu, Qi~Liu, and Lingpeng Kong.
\newblock Retrieved sequence augmentation for protein representation learning.
\newblock \emph{bioRxiv}, pp.\  2023--02, 2023.

\bibitem[Madani et~al.(2020)Madani, McCann, Naik, Keskar, Anand, Eguchi, Huang, and Socher]{madani2020progen}
Ali Madani, Bryan McCann, Nikhil Naik, Nitish~Shirish Keskar, Namrata Anand, Raphael~R Eguchi, Po-Ssu Huang, and Richard Socher.
\newblock Progen: Language modeling for protein generation.
\newblock \emph{arXiv preprint arXiv:2004.03497}, 2020.

\bibitem[Pal et~al.(2022)Pal, Umapathi, and Sankarasubbu]{pal2022medmcqa}
Ankit Pal, Logesh~Kumar Umapathi, and Malaikannan Sankarasubbu.
\newblock Medmcqa: A large-scale multi-subject multi-choice dataset for medical domain question answering.
\newblock In \emph{Conference on health, inference, and learning}, pp.\  248--260. PMLR, 2022.

\bibitem[Park et~al.(2023)Park, Yoon, Luo, L{\'o}pez-Marrero, Johnstone, Yoo, and Alexander]{park2023comparative}
Gilchan Park, Byung-Jun Yoon, Xihaier Luo, Vanessa L{\'o}pez-Marrero, Patrick Johnstone, Shinjae Yoo, and Francis~J Alexander.
\newblock Comparative performance evaluation of large language models for extracting molecular interactions and pathway knowledge.
\newblock \emph{arXiv preprint arXiv:2307.08813}, 2023.

\bibitem[Park et~al.(2024)Park, Lee, and Kim]{park2024multi}
Junkil Park, Youhan Lee, and Jihan Kim.
\newblock Multi-modal conditioning for metal-organic frameworks generation using 3d modeling techniques.
\newblock 2024.

\bibitem[Roohani et~al.(2024)Roohani, Lee, Huang, Vora, Steinhart, Huang, Marson, Liang, and Leskovec]{roohani2024biodiscoveryagent}
Yusuf Roohani, Andrew Lee, Qian Huang, Jian Vora, Zachary Steinhart, Kexin Huang, Alexander Marson, Percy Liang, and Jure Leskovec.
\newblock Biodiscoveryagent: An ai agent for designing genetic perturbation experiments.
\newblock \emph{arXiv preprint arXiv:2405.17631}, 2024.

\bibitem[Shinn et~al.(2023)Shinn, Cassano, Gopinath, Narasimhan, and Yao]{shinn2023reflexion}
Noah Shinn, Federico Cassano, Ashwin Gopinath, Karthik~R Narasimhan, and Shunyu Yao.
\newblock Reflexion: Language agents with verbal reinforcement learning.
\newblock In \emph{Thirty-seventh Conference on Neural Information Processing Systems}, 2023.

\bibitem[Singhal et~al.(2023)Singhal, Azizi, Tu, Mahdavi, Wei, Chung, Scales, Tanwani, Cole-Lewis, Pfohl, et~al.]{singhal2023large}
Karan Singhal, Shekoofeh Azizi, Tao Tu, S~Sara Mahdavi, Jason Wei, Hyung~Won Chung, Nathan Scales, Ajay Tanwani, Heather Cole-Lewis, Stephen Pfohl, et~al.
\newblock Large language models encode clinical knowledge.
\newblock \emph{Nature}, 620\penalty0 (7972):\penalty0 172--180, 2023.

\bibitem[Sun et~al.(2023)Sun, Xu, Tang, Wang, Lin, Gong, Ni, Shum, and Guo]{sun2023think}
Jiashuo Sun, Chengjin Xu, Lumingyuan Tang, Saizhuo Wang, Chen Lin, Yeyun Gong, Lionel Ni, Heung-Yeung Shum, and Jian Guo.
\newblock Think-on-graph: Deep and responsible reasoning of large language model on knowledge graph.
\newblock In \emph{The Twelfth International Conference on Learning Representations}, 2023.

\bibitem[Tang et~al.(2024)Tang, Yang, Wei, Shi, Su, Cheng, Yin, and Huang]{tang2024graphgpt}
Jiabin Tang, Yuhao Yang, Wei Wei, Lei Shi, Lixin Su, Suqi Cheng, Dawei Yin, and Chao Huang.
\newblock Graphgpt: Graph instruction tuning for large language models.
\newblock In \emph{Proceedings of the 47th International ACM SIGIR Conference on Research and Development in Information Retrieval}, pp.\  491--500, 2024.

\bibitem[Valentini et~al.(2023)Valentini, Malchiodi, Gliozzo, Mesiti, Soto-Gomez, Cabri, Reese, Casiraghi, and Robinson]{valentini2023promises}
Giorgio Valentini, Dario Malchiodi, Jessica Gliozzo, Marco Mesiti, Mauricio Soto-Gomez, Alberto Cabri, Justin Reese, Elena Casiraghi, and Peter~N Robinson.
\newblock The promises of large language models for protein design and modeling.
\newblock \emph{Frontiers in Bioinformatics}, 3:\penalty0 1304099, 2023.

\bibitem[Vilares \& G{\'o}mez-Rodr{\'\i}guez(2019)Vilares and G{\'o}mez-Rodr{\'\i}guez]{vilares2019head}
David Vilares and Carlos G{\'o}mez-Rodr{\'\i}guez.
\newblock Head-qa: A healthcare dataset for complex reasoning.
\newblock \emph{arXiv preprint arXiv:1906.04701}, 2019.

\bibitem[Wang et~al.(2024)Wang, Feng, He, Tan, Han, and Tsvetkov]{wang2024can}
Heng Wang, Shangbin Feng, Tianxing He, Zhaoxuan Tan, Xiaochuang Han, and Yulia Tsvetkov.
\newblock Can language models solve graph problems in natural language?
\newblock \emph{Advances in Neural Information Processing Systems}, 36, 2024.

\bibitem[Wei et~al.(2022)Wei, Wang, Schuurmans, Bosma, Xia, Chi, Le, Zhou, et~al.]{wei2022chain}
Jason Wei, Xuezhi Wang, Dale Schuurmans, Maarten Bosma, Fei Xia, Ed~Chi, Quoc~V Le, Denny Zhou, et~al.
\newblock Chain-of-thought prompting elicits reasoning in large language models.
\newblock \emph{Advances in Neural Information Processing Systems}, 35:\penalty0 24824--24837, 2022.

\bibitem[Welbl et~al.(2018)Welbl, Stenetorp, and Riedel]{welbl2018constructing}
Johannes Welbl, Pontus Stenetorp, and Sebastian Riedel.
\newblock Constructing datasets for multi-hop reading comprehension across documents.
\newblock \emph{Transactions of the Association for Computational Linguistics}, 6:\penalty0 287--302, 2018.

\bibitem[Yao et~al.(2023)Yao, Zhao, Yu, Du, Shafran, Narasimhan, and Cao]{yao2023react}
Shunyu Yao, Jeffrey Zhao, Dian Yu, Nan Du, Izhak Shafran, Karthik Narasimhan, and Yuan Cao.
\newblock React: Synergizing reasoning and acting in language models, 2023.

\bibitem[Ye et~al.(2023)Ye, Zhang, Wang, Xu, Zhang, et~al.]{ye2023natural}
Ruosong Ye, Caiqi Zhang, Runhui Wang, Shuyuan Xu, Yongfeng Zhang, et~al.
\newblock Natural language is all a graph needs.
\newblock \emph{arXiv preprint arXiv:2308.07134}, 4\penalty0 (5):\penalty0 7, 2023.

\bibitem[Yu et~al.(2023)Yu, Jiang, Shi, Yu, Liu, Zhang, Kwok, Li, Weller, and Liu]{yu2023metamath}
Longhui Yu, Weisen Jiang, Han Shi, Jincheng Yu, Zhengying Liu, Yu~Zhang, James~T Kwok, Zhenguo Li, Adrian Weller, and Weiyang Liu.
\newblock Metamath: Bootstrap your own mathematical questions for large language models.
\newblock \emph{arXiv preprint arXiv:2309.12284}, 2023.

\bibitem[Zhao et~al.(2023)Zhao, Liu, Chang, Xu, Fu, Deng, Kong, and Liu]{zhao2023gimlet}
Haiteng Zhao, Shengchao Liu, Ma~Chang, Hannan Xu, Jie Fu, Zhihong Deng, Lingpeng Kong, and Qi~Liu.
\newblock Gimlet: A unified graph-text model for instruction-based molecule zero-shot learning.
\newblock \emph{Advances in Neural Information Processing Systems}, 36:\penalty0 5850--5887, 2023.

\bibitem[Zhao et~al.(2024)Zhao, Ma, Wang, Su, Kong, Xu, Deng, and Yang]{zhao2024empowering}
Haiteng Zhao, Chang Ma, Guoyin Wang, Jing Su, Lingpeng Kong, Jingjing Xu, Zhi-Hong Deng, and Hongxia Yang.
\newblock Empowering large language model agents through action learning.
\newblock \emph{arXiv preprint arXiv:2402.15809}, 2024.

\bibitem[Zheng et~al.(2023)Zheng, Alawadhi, Chheda, Neumann, Rampal, Liu, Nguyen, Lin, Rong, Siepmann, et~al.]{zheng2023shaping}
Zhiling Zheng, Ali~H Alawadhi, Saumil Chheda, S~Ephraim Neumann, Nakul Rampal, Shengchao Liu, Ha~L Nguyen, Yen-hsu Lin, Zichao Rong, J~Ilja Siepmann, et~al.
\newblock Shaping the water-harvesting behavior of metal--organic frameworks aided by fine-tuned gpt models.
\newblock \emph{Journal of the American Chemical Society}, 145\penalty0 (51):\penalty0 28284--28295, 2023.

\bibitem[Zhu et~al.(2022)Zhu, Shi, Zhang, Liu, Xu, Yuan, Zhang, Chen, Cai, Lu, et~al.]{zhu2022torchdrug}
Zhaocheng Zhu, Chence Shi, Zuobai Zhang, Shengchao Liu, Minghao Xu, Xinyu Yuan, Yangtian Zhang, Junkun Chen, Huiyu Cai, Jiarui Lu, et~al.
\newblock Torchdrug: A powerful and flexible machine learning platform for drug discovery.
\newblock \emph{arXiv preprint arXiv:2202.08320}, 2022.

\end{thebibliography}
\bibliographystyle{iclr2025_conference}

\newpage
\appendix
\section{Appendix}

\subsection{Dataset Case Illustration} \label{appendix data cases}

\textbf{Dimension 1: Inquiry Type}

\textbf{Category 1: Normal Source} 

Question: In the context of YopT-induced cleavage of Rho GTPases, is carboxyl methylation a necessary post-translational modification for the cysteine protease activity of YopT to occur?

Answer: No

Question: In U2OS-hGR osteosarcoma cells, does the glucocorticoid receptor (GR)-dependent mRNA expression of ladinin 1 require the presence of both Mediator subunits MED1 and MED14 for transcriptional regulation in response to dexamethasone treatment?

Answer: Yes

Question: In pancreatic acinar cells, How does the sensitivity to nicotinic acid adenine dinucleotide phosphate (NAADP) compare to the sensitivity to cyclic adenosine 5'-diphosphate-ribose (cADPR) and inositol trisphosphate (InsP3) in terms of triggering Ca2+ release?

Answer: Pancreatic acinar cells are significantly more sensitive to NAADP than to either cyclic adenosine 5'-diphosphate-ribose (cADPR) or inositol trisphosphate (InsP3).

Question: In the context of Mycobacterium tuberculosis signal transduction, what is the effect of TrcS-phosphate and calcium ions (Ca2+) on the transphosphorylation of the response regulator protein TrcR?

Answer: TrcS-phosphate and Ca2+ are required for the transphosphorylation of TrcR.

\textbf{Category 2: Perturbed Source} 

Question: In the context of human papillomavirus type 16 (HPV16) E7 gene product's role in cellular transformation, can a transcriptionally inactive c-Jun deletion mutant that retains E7 binding capability interfere with the E7-induced transformation of rat embryo fibroblasts when co-expressed with an activated ras oncogene?

Answer: Yes

Question: In HTLV-1-infected T-cell lines, does the application of short interfering RNA (siRNA) targeting JunD result in an increase in matrix metalloproteinase-7 (MMP-7) mRNA expression?

Answer: No

Question: In the context of opioid withdrawal, how does blocking A1-adenosine receptors with 8-cyclopentyl-1, 3-dipropylxantine affect the response to mu-opioid receptor activation on the amplitude of GABAB-mediated IPSPs in slices taken from morphine-treated guinea pigs?

Answer: Blocking A1-adenosine receptors with 8-cyclopentyl-1, 3-dipropylxantine allows mu-opioid receptor activation to inhibit the IPSP at all concentrations and increases the maximal inhibition in withdrawn slices.

Question: In 3T3-L1 adipocytes, What is the effect of platelet-derived growth factor treatment on the tyrosine phosphorylation of APS?

Answer: Platelet-derived growth factor treatment results in no APS phosphorylation.

\textbf{Dimension 2: Extra Condition}

\textbf{Category 1: Natural Condition}

Question: In the context of Kaposi's sarcoma-associated herpesvirus (KSHV) infection, can viral interleukin-6 (vIL-6) induce the up-regulation of DNA methyltransferase 1 (DNMT1) in endothelial cells independently of signal transducer and activator of transcription 3 (STAT3) activation?

Answer: No

Question: In BJAB cells, does the expression of the K5 protein, encoded by the Kaposi's sarcoma-associated herpesvirus (KSHV), lead to a reduction in the surface expression of intercellular adhesion molecule 1 (ICAM-1)?

Answer: Yes

Question: In human hepatoma HuH-7 cells, What is the effect of transient expression of the Hepatitis C virus (HCV) core protein on the trans-activation ability of the cellular putative RNA helicase CAP-Rf?

Answer: Transient expression of HCV core protein in human hepatoma HuH-7 cells potentiates the trans-activation effect of CAP-Rf on gene expression.

Question: In Caenorhabditis elegans, What is the effect of being heterozygous for the cet-1 gene on body length compared to wild-type and null mutant individuals?

Answer: Heterozygotes for cet-1 display body lengths ranging between null mutant and wild type.

\textbf{Category 2: Intervened Condition} 

Question: In the Edar signalling pathway, can NF-kappaB activation still be induced by Edar-associated death domain (Edaradd) in the presence of dominant negative forms of TNF-receptor-associated factor 6 (TRAF6)?

Answer: No

Question: In mice treated with the angiotensin-converting enzyme inhibitor ramipril, do those lacking the specific PACAP receptor (PAC1-/-) exhibit lower plasma renin concentrations compared to their wild-type littermates?

Answer: Yes

Question: In human lung cancer cell lines with methylated retinoic acid receptor-beta (RARbeta) P2 promoter, What is the effect of demethylation by 5-aza-2'-deoxycytidine on the expression of RARbeta2 and RARbeta4 isoforms?

Answer: Demethylation by exposure to 5-aza-2'-deoxycytidine restores the expression of RARbeta2 and RARbeta4 in methylated tumor lines.

Question: In the context of hematopoietic progenitor cell proliferation, What is the effect of Flt3 ligand (FL) on progenitor cells from mice deficient in the signal transducer and activator of transcription 5a (Stat5a)?

Answer: Flt3 ligand (FL) does not act on progenitors from marrows of Stat5a(-/-) mice.

\textbf{Dimension 3: Investigation Target}

\textbf{Category 1: Single Component as Target} 

Question: In MELN cells, which are derived from MCF-7 breast cancer cells and stably express estrogen receptor alpha (ERalpha), does exposure to mono-n-butyl ester phthalic acid result in an increase in vascular endothelial growth factor (VEGF) secretion?

Answer: No

Question: In the context of non-alcoholic fatty liver disease (NAFLD) induced by a choline-deficient, ethionine-supplemented (CDE) diet in C57BL/6J mice, is the protein level of peroxisome-proliferator-activated receptor-gamma coactivator 1alpha (PGC1alpha) decreased in comparison to control livers?

Answer: Yes

Question: In the context of Mycobacterium tuberculosis signal transduction, what is the effect of TrcS-phosphate and calcium ions (Ca2+) on the transphosphorylation of the response regulator protein TrcR?

Answer: TrcS-phosphate and Ca2+ are required for the transphosphorylation of TrcR.

Question: In the context of myeloma cells, What is the effect of elevated heparanase expression on the expression levels of urokinase-type plasminogen activator (uPA) and the uPA receptor?

Answer: Elevation in heparanase expression in myeloma cells increases urokinase-type plasminogen activator (uPA) and uPA receptor expression levels.

\textbf{Category 2: Components Interaction as Target} 

Question: In the context of growth hormone-releasing hormone (GHRH) stimulation of growth hormone (GH) gene expression in anterior pituitary somatotrophs, does the CBP-Pit-1 pathway, which involves the interaction between CREB binding protein (CBP) and the pituitary-specific transcription factor Pit-1, require the presence of cAMP-response element binding protein (CREB) to mediate its effects on the human GH promoter?

Answer: No

Question: In the context of mitophagy, does the ubiquitin-binding protein RABGEF1 play a role in recruiting the downstream Rab GTPases, RAB5 and RAB7A, to damaged mitochondria that have been ubiquitinated by Parkin?

Answer: Yes

Question: In human lung cancer cells treated with the nitrosamine 4-(methylnitrosamino)-1-(3-pyridyl)-1-butanone (NNK), What is the role of Bcl2 phosphorylation at Ser(70) in the interaction between Bcl2 and c-Myc oncogenic proteins?

Answer: Phosphorylation of Bcl2 at Ser(70) promotes a direct interaction between Bcl2 and c-Myc, enhancing the half-life of the c-Myc protein.

Question: In cells infected with Epstein-Barr virus (EBV), How does the BGLF4 protein kinase affect the recruitment of active interferon regulatory factor 3 (IRF3) to the IRF3-responsive element containing the interferon-beta (IFN-beta) promoter region?

Answer: BGLF4 reduces the amount of active IRF3 recruited to the IRF3-responsive element containing the IFN-beta promoter region.

\textbf{Category 3: Function as Target}

Question: In human hepatoma cells (Huh-7) and normal human hepatocytes (Hc) infected with an adenovirus expressing a mutated form of IkappaBalpha (Ad5IkappaB), does pretreatment with N,N-dimethylsphingosine (DMS), an inhibitor of sphingosine kinase (SphK), lead to an increase in the number of apoptotic cells induced by tumor necrosis factor-alpha (TNF-alpha) stimulation?

Answer: Yes

Question: In Escherichia coli, do mutants lacking acyl-acyl carrier protein (acyl-ACP) synthetase activity (aas mutants) retain the ability to incorporate exogenous fatty acids into the major outer membrane lipoprotein through an acyl-CoA-independent pathway?

Answer: No

Question: In the context of myocytes expressing adiponectin receptor 1 (AdipoR1), What is the effect of adiponectin on extracellular calcium (Ca2+) influx?

Answer: Adiponectin induces extracellular Ca(2+) influx by adiponectin receptor 1 (AdipoR1) in myocytes.

Question: In the context of lung inflammation resolution, What is the role of cholesterol 25-hydroxylase (Ch25h) in alveolar macrophages (AM) during the resolution phase of inflammation?

Answer: Ch25h is induced in macrophages upon their encounter with apoptotic cells and is required for LXR-dependent prevention of AM lipid overload, induction of Mertk, efferocytic resolution of airway neutrophilia, and induction of TGF-beta.

\subsection{Data Creation and Filter Pipeline} \label{appendix: data creation}

\begin{figure*}[!h]
    \centering    
    \renewcommand{\thesubfigure}{} 
    \subfigure[]{\includegraphics[width=0.95\linewidth]{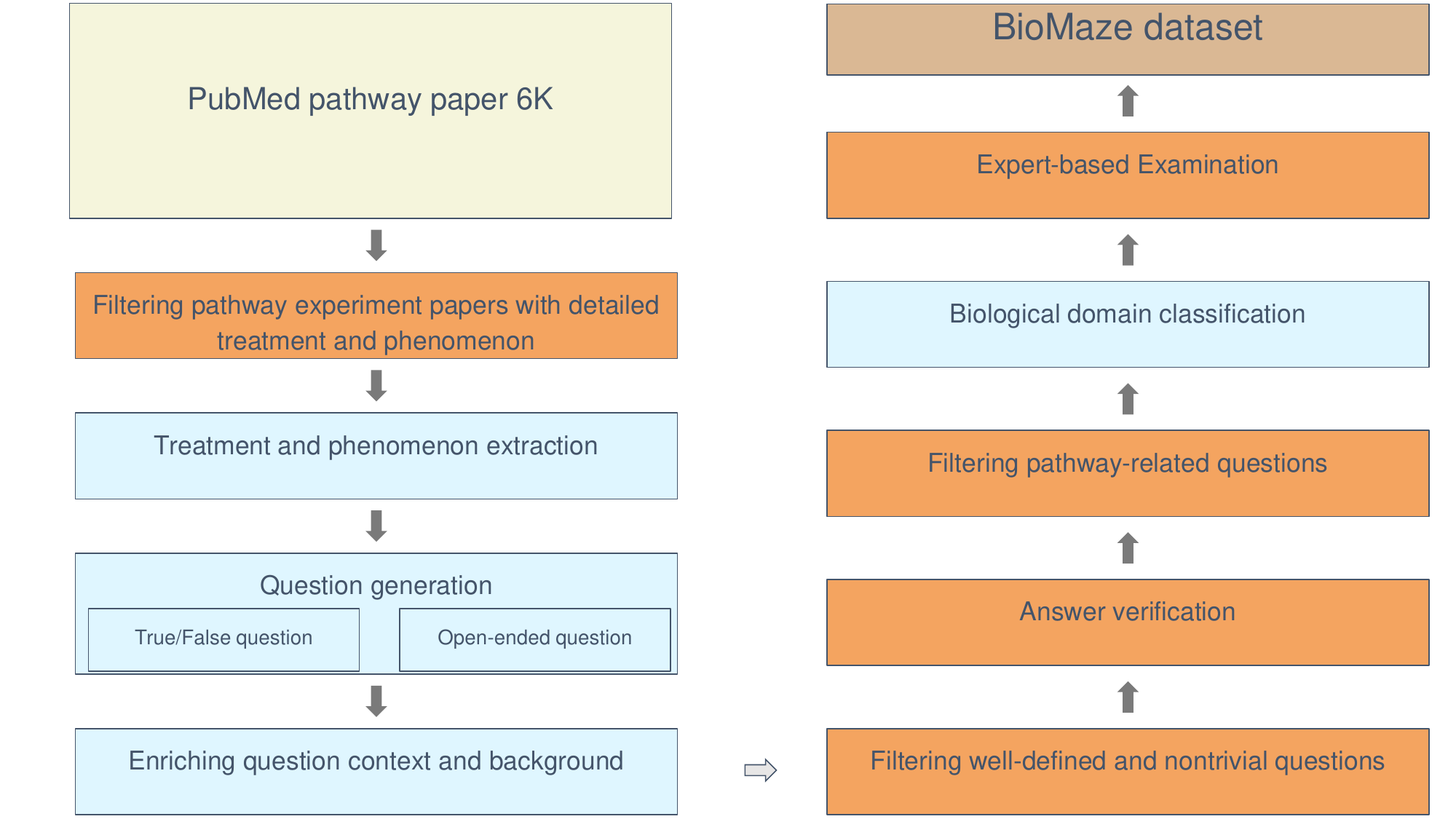}}
    \caption{Dataset \benchname creation pipeline.}
    \label{dataset_pipeline}
\end{figure*}

The overall dataset creation pipeline is shown in Figure \ref{dataset_pipeline}. 

To ensure question quality, we employ a two-step process. First, we create and filter questions using an advanced language model ( GPT-4 and
LLaMa 3.1-405B) to assess their relevance and clarity. Subsequently, each question undergoes a final quality check by human experts.

The well-define filter removes questions that are poorly defined, unpredictable (e.g., asking for specific measurement values), or require more than one prediction, and the nontrivial filter removes the data that answers revealed in the question's context. 

To validate the answer quality, we require the LLM to answer the questions based on the original paper's content. The model is explicitly instructed to respond with Undetermined if it cannot confidently generate
an
answer. Each question is tested five times, and only questions that are consistently answered correctly (i.e., aligned
with the intended label) and not marked as Undetermined in any of the trials are retained. This process helps eliminate
questions with incorrect labels, ambiguous phrasing, or poor structure.

In the final stage, human experts perform the quality check to further refine the questions, excluding data that are poorly defined, overly complex (e.g., asking for multiple facts), incorrectly labeled, or of any other low quality, ensuring the dataset's overall reliability and usability. Approximately 60\% of the data is filtered out at this stage.

\subsection{Question key words Distribution}

\begin{figure*}[!h]
    \centering    
    \renewcommand{\thesubfigure}{} 
    \subfigure[]{\includegraphics[width=0.48\linewidth]{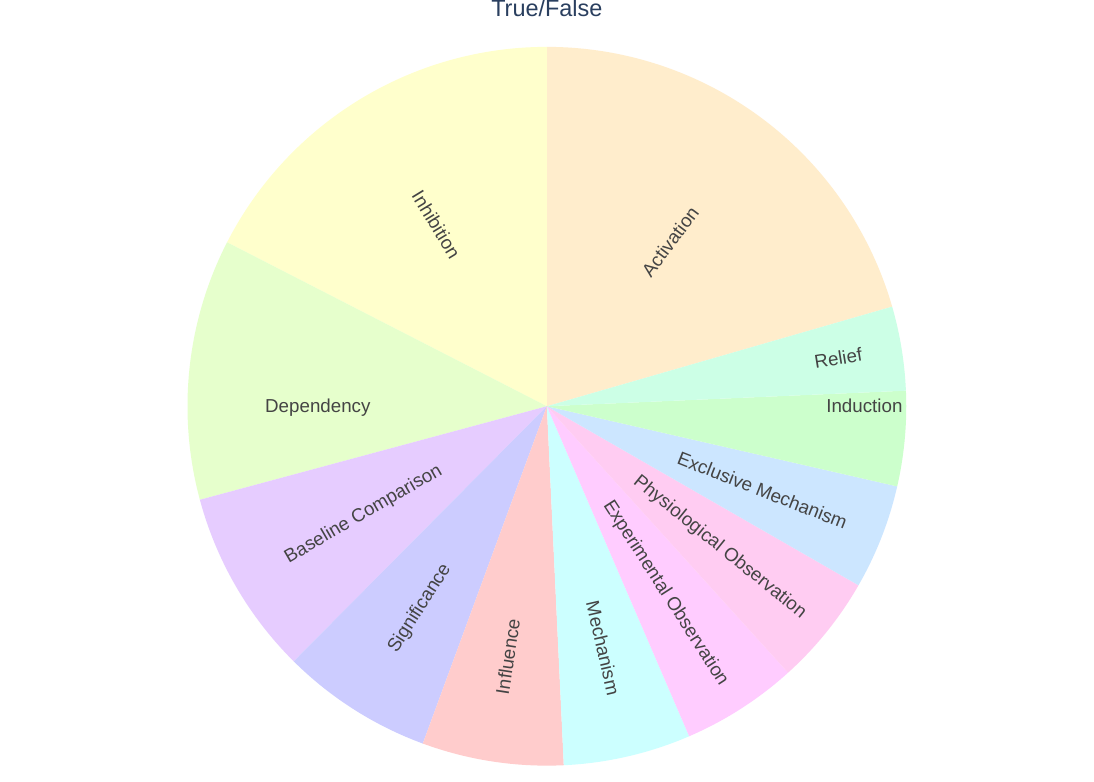}} 
    \subfigure[]{\includegraphics[width=0.48\linewidth]{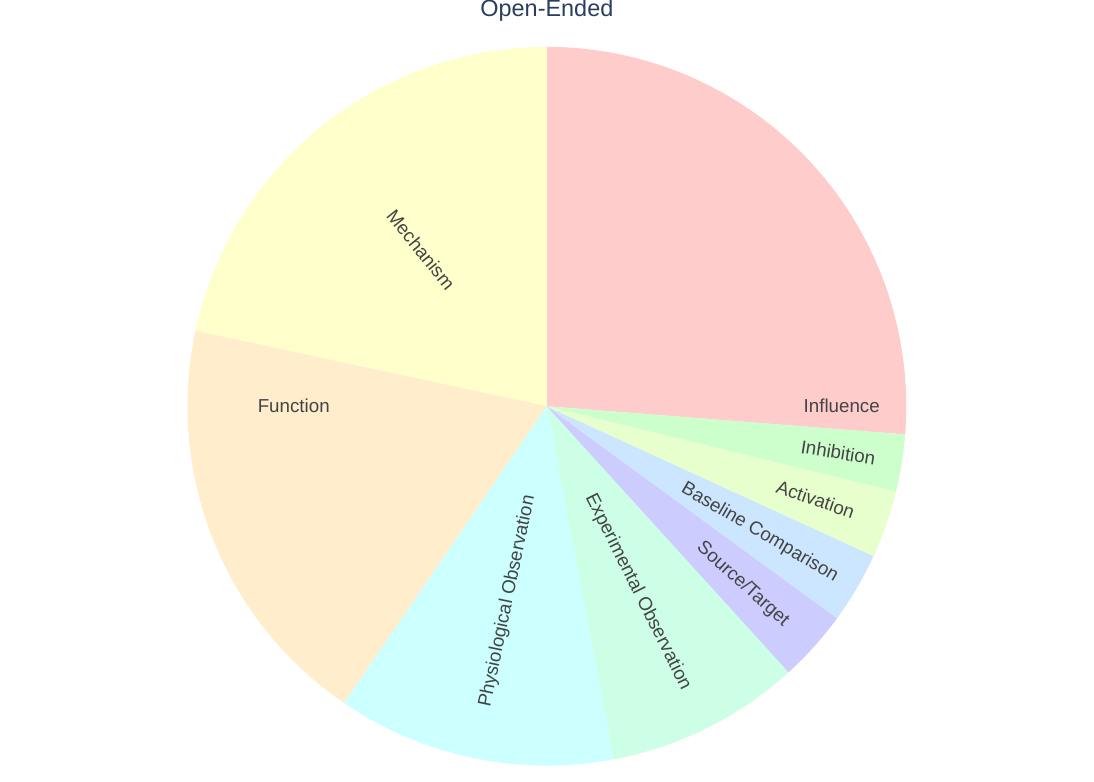}} 
    \caption{Dataset key words distribution.}
    \label{fig key words fan}
\end{figure*}

We present the distribution of question keywords in Figure \ref{fig key words fan}. While these keywords do not directly correspond to the three main categories we primarily use, they offer an additional perspective on the dataset. Below are explanations of the keywords:

General Influence Inquiry: Can x influence y or not?

Activation Inquiry: Can x activate y?

Inhibition Inquiry: Can x inhibit y?

Dependency Inquiry: Is y dependent on x?

Induction Question: Can x induce y?

Relief Inquiry: Can x relieve y?

Mechanism Question: Does x influence y via a specific mechanism?

Exclusive Mechanism Question: Is a specific mechanism the only mechanism for process z?

Significance Inquiry: Will x cause a significant/insignificant phenomenon y?

Baseline Comparison Question: Is x different from the baseline?

Experimental Observation Question: Will a specific detailed phenomenon be observed in the experiment?

Physiological Observation Question: Will a specific phenomenon be observed in the body?

\subsection{Pathway Graph Database Statistics} \label{appendix pathway statistic}

\begin{table}[!h]
\centering
\caption{Data statistic of our pathway network database.}
\label{table pathway statistic}
\begin{tabular}{ccc}
\toprule
Entries & Edges & Involved biological process \\
8939    & 15131 & 2265                        \\
\bottomrule
\end{tabular}
\end{table}

\subsection{Implementation of Subgraph Retriever} \label{appendix subgraph retriever}

Since the connectivity of the pathway graph is crucial for enhancing reasoning in biological systems, we developed the retrieval API designed to find the optimal connected subgraph $S^*=\operatorname{Search\_Subgraph}(\texttt{query}, N)$, where $S^*$ is the retrieved subgraph, $\texttt{query}$ represents the query content, such as keywords, and $N$ is the target size. The goal is to match a given target size as closely as possible while maximizing the matching score. This is formulated as a optimization problem:
\begin{equation}
\begin{aligned}
S^*&=\operatorname{Search\_Subgraph}(\texttt{query}, N)\\&= \underset{\substack{S \subseteq P, \\ |S|=N,\\S \text { is connected}}}{\operatorname{argmax}} \sum_{i \in V_S \cup E_S} \operatorname{score}(i, \texttt{query})\\
\end{aligned}
\end{equation}
Here, the overall pathway network is denoted by $P$, with $V_S$ and $E_S$ representing the node and edge sets of subgraph $S$. 



The optimization problem is hard to solve directly for huge graph databse. Inspire by \cite{he2024g} that convert optimal subgraph problem into a Prize-Collecting Steiner Tree (PCST) problem, we solve the problem as a bi-level optimization 
as follows:

\begin{equation}
\small
\begin{aligned}
&S(C_e)=\underset{\substack{S \subseteq P, \\ S \text { is connected }}}{\operatorname{argmax}} \sum_{i \in V_S \cup E_S} \operatorname{score}(i, \texttt{query})-\left|E_S\right| \times C_e\\
& C_e^* = \underset{\substack{C_e}}{\operatorname{argmin}}  | |S(C_e)|-N|\\
\end{aligned}
\end{equation}

Then 
$$
\operatorname{Search\_Subgraph}(\texttt{query}, N)= S(C_e^*)\\
$$

The inner optimization problem, a Prize-Collecting Steiner Tree (PCST) problem, seeks to identify a connected subgraph that maximizes the total score of its nodes and edges while minimizing the edge costs, as utilized in \citep{he2024g}. The outer optimization problem minimizes the difference between the optimal subgraph size and the target size. In practice, we apply a fast-solving method for the inner problem proposed by \citep{hegde2015nearly} and employ a binary root search algorithm for the outer problem.

\textbf{Graph-to-Text Encoder} For retrieved results which is a list of triples (e.g., the output of $\operatorname{Search\_Triple}$ or $\operatorname{Search\_Subgraph}$), the result $S$ is encoded into text $o$ using the following process:
\begin{equation}
\begin{aligned}
o&=\operatorname{TripleToText}(\operatorname{DFSOrder}(S))
\end{aligned}
\end{equation}
Function $\operatorname{DFSOrder}$ arranges the triples in depth-first search (DFS) order. Unlike other methods, such as relevance scoring, DFS order more closely aligns with the reasoning process through the subgraph, reflecting the natural progression of biological functions. The ordered subgraph $\hat{S}$ is converted into text format for the LLM by encoding each triple as a string in the following format:
$\text{ Head } |  \text{ Tail } | \text{ Relation and Biological Process}$.


\subsection{\modelname Case} \label{appendix agent case}

\begin{lstlisting}[style=QuestionType]
Please explore pathways to find relevant information regarding the following question: Worldwide, one of the most prevalent forms of chronic disease is alcoholic fatty liver, which may progress to more severe forms of liver injury. Chronic ethanol exposure is thought to cause enhanced hepatic lipogenesis and impaired fatty acid oxidation. Adiponectin is an adipose-derived hormone that has been associated with the development of alcoholic liver steatosis in rodent models. Does adiponectin have a protective role against alcoholic liver steatosis?
\end{lstlisting}
\begin{lstlisting}[style=AnswerType]
Action: Thought: To find information regarding the role of adiponectin in protecting against alcoholic liver steatosis, I can use the search_biopathway_subgraph_global tool to explore relevant biological pathways. I should convert all the question content into keywords to enhance retrieval quality. Action:
     search_biopathway_subgraph_global(['Adiponectin, adipose-derived hormone, associated with the development of alcoholic liver steatosis in rodent models, protective role against alcoholic liver steatosis, chronic ethanol exposure, enhanced hepatic lipogenesis, impaired fatty acid oxidation, liver injury'])
     End Action
\end{lstlisting}
\begin{lstlisting}[style=QuestionType]
State: 0) C00469: Ethanol; Ethyl alcohol; Methylcarbinol | 406999: microRNA 217 MIR217 MIRN217 mir-217 | PCrel indirect effect activation | hsa04936: Alcoholic liver disease
1) 406999: microRNA 217 MIR217 MIRN217 mir-217 | 23411: sirtuin 1 SIRT1 SIR2 SIR2L1 SIR2alpha | PPrel inhibition indirect effect | hsa04936: Alcoholic liver disease
2) 406999: microRNA 217 MIR217 MIRN217 mir-217 | 29896 6434: TRA2A AWMS1 HSU53209 | PPrel inhibition indirect effect | hsa04936: Alcoholic liver disease
3) 29896 6434: TRA2A AWMS1 HSU53209 | 23175 64900 9663: LPIN1 PAP1 | PPrel activation | hsa04936: Alcoholic liver disease
4) 23175 64900 9663: LPIN1 PAP1 | 6720: sterol regulatory element binding transcription factor 1 SREBF1 HMD IFAP2 SREBP1 bHLHd1 | PPrel inhibition | hsa04936: Alcoholic liver disease
5) 6720: sterol regulatory element binding transcription factor 1 SREBF1 HMD IFAP2 SREBP1 bHLHd1 | 6319 79966: SCD FADS5 MSTP008 SCD1 SCDOS hSCD1 | GErel expression | hsa04152: AMPK signaling pathway hsa04936: Alcoholic liver disease
6) 6720: sterol regulatory element binding transcription factor 1 SREBF1 HMD IFAP2 SREBP1 bHLHd1 | 23175 64900 9663: LPIN1 PAP1 | GErel expression | hsa04936: Alcoholic liver disease
7) C00469: Ethanol; Ethyl alcohol; Methylcarbinol | 51094 79602: ADIPOR1 ACDCR1 CGI-45 CGI45 PAQR1 TESBP1A | PCrel inhibition indirect effect | hsa04936: Alcoholic liver disease
8) C00469: Ethanol; Ethyl alcohol; Methylcarbinol | 51422 53632 5562 5563 5564 5565 5571: PRKAG2 AAKG AAKG2 CMH6 H91620p WPWS | PCrel inhibition indirect effect | hsa04936: Alcoholic liver disease
9) C00469: Ethanol; Ethyl alcohol; Methylcarbinol | 50507: NADPH oxidase 4 NOX4 KOX KOX-1 RENOX | PCrel activation | hsa04936: Alcoholic liver disease
10) 50507: NADPH oxidase 4 NOX4 KOX KOX-1 RENOX | C22381: Reactive oxygen species; ROS | PCrel activation | hsa04936: Alcoholic liver disease
11) C22381: Reactive oxygen species; ROS | 51422 53632 5562 5563 5564 5565 5571: PRKAG2 AAKG AAKG2 CMH6 H91620p WPWS | PCrel inhibition indirect effect | hsa04936: Alcoholic liver disease
12) C22381: Reactive oxygen species; ROS | 10000 207 208: AKT3 MPPH MPPH2 PKB-GAMMA PKBG PRKBG RAC-PK-gamma RAC-gamma STK-2 | PCrel inhibition | hsa04936: Alcoholic liver disease
13) 10000 207 208: AKT3 MPPH MPPH2 PKB-GAMMA PKBG PRKBG RAC-PK-gamma RAC-gamma STK-2 | 1499: catenin beta 1 CTNNB1 CTNNB EVR7 MRD19 NEDSDV armadillo | PPrel activation phosphorylation | hsa04936: Alcoholic liver disease
14) C22381: Reactive oxygen species; ROS | 5606 5608: MAP2K3 MAPKK3 MEK3 MKK3 PRKMK3 SAPKK-2 SAPKK2 | PCrel activation | hsa04936: Alcoholic liver disease
15) C00469: Ethanol; Ethyl alcohol; Methylcarbinol | 712 713 714: C1QA C1QD1 | PCrel activation | hsa04936: Alcoholic liver disease
16) C00469: Ethanol; Ethyl alcohol; Methylcarbinol | 3605: interleukin 17A IL17A CTLA-8 CTLA8 IL-17 IL-17A IL17 ILA17 | PCrel indirect effect activation | hsa04936: Alcoholic liver disease
17) C00084: Acetaldehyde; Ethanal | 51422 53632 5562 5563 5564 5565 5571: PRKAG2 AAKG AAKG2 CMH6 H91620p WPWS | PCrel inhibition indirect effect | hsa04936: Alcoholic liver disease
18) 9370: adiponectin C1Q and collagen domain containing ADIPOQ ACDC ACRP30 ADIPQTL1 ADPN APM-1 APM1 GBP28 | 51094 79602: ADIPOR1 ACDCR1 CGI-45 CGI45 PAQR1 TESBP1A | PPrel activate activation | hsa04152: AMPK signaling pathway hsa04211: Longevity regulating pathway hsa04936: Alcoholic liver disease hsa04920: Adipocytokine signaling pathway N01788: ADIPOQ-ADIPOR signaling pathway
19) 7132: TNF receptor superfamily member 1A TNFRSF1A CD120a FPF TBP1 TNF-R TNF-R-I TNF-R55 TNFAR TNFR1 TNFR55 TNFR60 p55 p55-R p60 | 6720: sterol regulatory element binding transcription factor 1 SREBF1 HMD IFAP2 SREBP1 bHLHd1 | PPrel indirect effect activation | hsa04936: Alcoholic liver disease
\end{lstlisting}
\begin{lstlisting}[style=AnswerType]
Action: Thought: I have gathered a large number of pathways related to the question. To better understand the role of adiponectin in protecting against alcoholic liver steatosis, I will check the neighboring pathways of Pathway 18, which is directly related to adiponectin. I will use the function search_biopathway_triple_N_hop_subgraph with history_line_id set to 18. The keywords are the same as previously used.
     Action:
     search_biopathway_triple_N_hop_subgraph(18, ['Adiponectin, adipose-derived hormone, associated with the development of alcoholic liver steatosis in rodent models, protective role against alcoholic liver steatosis, chronic ethanol exposure, enhanced hepatic lipogenesis, impaired fatty acid oxidation, liver injury'])
     End Action
\end{lstlisting}
\begin{lstlisting}[style=QuestionType]
State: 20) C00219: Arachidonate; Arachidonic acid; (5Z8Z11Z14Z)-Icosatetraenoic acid; cis-581114-Eicosatetraenoic acid; 5Z8Z11Z14Z-Eicosatetraenoic acid; | 2180 2181 2182 23205 23305 51703: ACSL1 ACS1 FACL1 FACL2 LACS LACS1 LACS2 | pathway bind to | hsa04216: Ferroptosis N01590: Arachidonate/Adrenic acid metabolism
21) 2180 2181 2182 23205 23305 51703 81616: ACSL1 ACS1 FACL1 FACL2 LACS LACS1 LACS2 | C00162: Fatty acid | PCrel  | hsa04920: Adipocytokine signaling pathway
22) 1376: carnitine palmitoyltransferase 2 CPT2 CPT1 CPTASE IIAE4 | 51 8310: ACOX1 ACOX AOX MITCH PALMCOX SCOX | ECrel compound | hsa00071: Fatty acid degradation
23) 1376: carnitine palmitoyltransferase 2 CPT2 CPT1 CPTASE IIAE4 | 126129 1374 1375: CPT1C CATL1 CPT1-B CPT1P CPTI-B CPTIC SPG73 | ECrel compound | hsa00071: Fatty acid degradation
24) 126129 1374 1375: CPT1C CATL1 CPT1-B CPT1P CPTI-B CPTIC SPG73 | 2180 2181 2182 23205 23305 51703 81616: ACSL1 ACS1 FACL1 FACL2 LACS LACS1 LACS2 | ECrel compound | hsa00071: Fatty acid degradation hsa04920: Adipocytokine signaling pathway
25) 4217: mitogen-activated protein kinase kinase kinase 5 MAP3K5 ASK1 MAPKKK5 MEKK5 | 5609 6416: MAP2K7 JNKK2 MAPKK7 MEK MEK_7 MKK7 PRKMK7 SAPKK-4 SAPKK4 | PPrel activate activation phosphorylation | hsa05418: Fluid shear stress and atherosclerosis hsa04668: TNF signaling pathway hsa04936: Alcoholic liver disease hsa05208: Chemical carcinogenesis - reactive oxygen species N01407: Metals to JNK signaling pathway
26) 5609 6416: MAP2K7 JNKK2 MAPKK7 MEK MEK_7 MKK7 PRKMK7 SAPKK-4 SAPKK4 | 5599 5601 5602: MAPK8 JNK JNK-46 JNK1 JNK1A2 JNK21B1/2 PRKM8 | PPrel activate activation phosphorylation | hsa05418: Fluid shear stress and atherosclerosis hsa05135: Yersinia infection hsa05417: Lipid and atherosclerosis hsa05167: Kaposi sarcoma-associated herpesvirus infection hsa04620: Toll-like receptor signaling pathway hsa04668: TNF signaling pathway
27) 5599 5601 5602: MAPK8 JNK JNK-46 JNK1 JNK1A2 JNK21B1/2 PRKM8 SAPK1 SAPK1c | 3667 8471 8660: IRS1 HIRS-1 | PPrel inhibition phosphorylation | hsa04910: Insulin signaling pathway hsa05010: Alzheimer disease hsa04930: Type II diabetes mellitus hsa04920: Adipocytokine signaling pathway
28) 5599 5601 5602: MAPK8 JNK JNK-46 JNK1 JNK1A2 JNK21B1/2 PRKM8 SAPK1 SAPK1c | C00162: Fatty acid | PCrel  | hsa04930: Type II diabetes mellitus
29) 4790 5970: NFKB1 CVID12 EBP-1 KBF1 NF-kB NF-kB1 NF-kappa-B1 NF-kappaB NF-kappabeta NFKB-p105 | 2919 2920 2921: CXCL1 FSP GRO1 GROa MGSA MGSA-a NAP-3 SCYB1 | PPrel activation expression | hsa04621: NOD-like receptor signaling pathway hsa05167: Kaposi sarcoma-associated herpesvirus infection hsa05417: Lipid and atherosclerosis hsa05120: Epithelial cell signaling in Helicobacter pylori infection hsa04936: Alcoholic liver disease
30) 4790 5970: NFKB1 CVID12 EBP-1 KBF1 NF-kB NF-kB1 NF-kappa-B1 NF-kappaB NF-kappabeta NFKB-p105 NFKB-p50 | 4792: NFKB inhibitor alpha NFKBIA EDAID2 IKBA MAD-3 NFKBI | PPrel missing interaction dissociation | hsa05215: Prostate cancer hsa05167: Kaposi sarcoma-associated herpesvirus infection hsa05161: Hepatitis B hsa05220: Chronic myeloid leukemia hsa05160: Hepatitis C hsa04936: Alcoholic liver disease hsa04657: IL-17 signaling pathway
31) 4790 5970: NFKB1 CVID12 EBP-1 KBF1 NF-kB NF-kB1 NF-kappa-B1 NF-kappaB NF-kappabeta NFKB-p105 NFKB-p50 NFkappaB | 9021: suppressor of cytokine signaling 3 SOCS3 ATOD4 CIS3 Cish3 SOCS-3 SSI-3 SSI3 | GErel expression | hsa04920: Adipocytokine signaling pathway
32) 9021: suppressor of cytokine signaling 3 SOCS3 ATOD4 CIS3 Cish3 SOCS-3 SSI-3 SSI3 | 3667 8471 8660: IRS1 HIRS-1 | PPrel inhibition | hsa04920: Adipocytokine signaling pathway
33) 51094 79602: ADIPOR1 ACDCR1 CGI-45 CGI45 PAQR1 TESBP1A | C00076: Calcium cation; Ca2+; Calcium(2+); Calcium ion | PCrel indirect effect activation | hsa04152: AMPK signaling pathway hsa04936: Alcoholic liver disease
34) C00076: Calcium cation; Ca2+; Calcium(2+); Calcium ion | 10645: calcium/calmodulin dependent protein kinase kinase 2 CAMKK2 CAMKK CAMKKB | PCrel activation | hsa04152: AMPK signaling pathway hsa04211: Longevity regulating pathway hsa04936: Alcoholic liver disease hsa04920: Adipocytokine signaling pathway
35) 10645: calcium/calmodulin dependent protein kinase kinase 2 CAMKK2 CAMKK CAMKKB | 51422 53632 5562 5563 5564 5565 5571: PRKAG2 AAKG AAKG2 CMH6 H91620p WPWS | PPrel activation phosphorylation | hsa04152: AMPK signaling pathway hsa04921: Oxytocin signaling pathway hsa04211: Longevity regulating pathway hsa04936: Alcoholic liver disease hsa04920: Adipocytokine signaling pathway
36) 51094 79602: ADIPOR1 ACDCR1 CGI-45 CGI45 PAQR1 TESBP1A | 5465: peroxisome proliferator activated receptor alpha PPARA NR1C1 PPAR PPAR-alpha PPARalpha hPPAR | PPrel indirect effect activation | hsa04920: Adipocytokine signaling pathway
37) 5465: peroxisome proliferator activated receptor alpha PPARA NR1C1 PPAR PPAR-alpha PPARalpha hPPAR | 126129 1374 1375: CPT1C CATL1 CPT1-B CPT1P CPTI-B CPTIC SPG73 | GErel expression | hsa04920: Adipocytokine signaling pathway
38) 5465: peroxisome proliferator activated receptor alpha PPARA NR1C1 PPAR PPAR-alpha PPARalpha hPPAR | C00638: Long-chain fatty acid; Higher fatty acid | PCrel  | hsa04920: Adipocytokine signaling pathway
\end{lstlisting}
\begin{lstlisting}[style=AnswerType]
Action: Thought: I have gathered a large number of pathways related to the question. To better understand the role of adiponectin in protecting against alcoholic liver steatosis, I will check the neighboring pathways of Pathway 18, which is directly related to adiponectin. I will use the function search_biopathway_triple_N_hop_subgraph with history_line_id set to 18. The keywords are the same as previously used.
     Action:
     search_biopathway_triple_N_hop_subgraph(18, ['Adiponectin, adipose-derived hormone, associated with the development of alcoholic liver steatosis in rodent models, protective role against alcoholic liver steatosis, chronic ethanol exposure, enhanced hepatic lipogenesis, impaired fatty acid oxidation, liver injury'])
     End Action
\end{lstlisting}
\begin{lstlisting}[style=QuestionType]
State: No new pathways were found besides those previously seen.
\end{lstlisting}
\begin{lstlisting}[style=AnswerType]
Action: Thought: Since no new pathways were found, I will finish the exploration.
Answer:
Finished.
\end{lstlisting}
\begin{lstlisting}[style=QuestionType]
State: You finished the task.
\end{lstlisting}

\subsection{Baseline Details} \label{appendix baseline}

\textbf{Chain-of-Thought (CoT)} \citep{wei2022chain,kojima2022large} prompting encourages language models to generate step-by-step reasoning to solve tasks. We adopt CoT as the unaugmented standard method for LLM reasoning.

\textbf{Chain-of-Knowledge (CoK)} \citep{li2023chain} enhances chain-of-thought reasoning through knowledge verification, enabling language models to handle knowledge-intensive tasks. CoK first generates reasoning using chain-of-thought processes, then employs knowledge triples to verify the accuracy of the reasoning. While the reasoning of CoK is primarily driven by the language model, graph-based information is used for fact verification.

\textbf{Think-of-Graph (ToG)} \citep{sun2023think} is an interactive reasoning method designed to actively navigate knowledge graphs for question solving. It primarily uses large language models to prune knowledge graph edges, thereby enabling efficient knowledge acquisition from complex graphs. The reasoning process in ToG is guided by graph navigation.

\textbf{G-Retriever} \citep{he2024g} is a graph retriever-augmented generation method that retrieves relevant subgraphs from a database and generates answers based on the retrieved subgraphs. While the original model in their work uses a graph encoder to encode graph data as a separate modality, in this work, we directly implement the graph-to-text encoder for improved versatility and better comparability with other methods.

\subsection{{Failure Reason Category Cases}}

We classify the error reason of biological pathway reasoning into four types:
(1) \textbf{Unresolved Conclusion} For cases where the model fails to provide a definitive answer, indicating uncertainty or belief that the answer is unknown.
(2) \textbf{Incomplete Answer}  When the response lacks essential details, such as missing the requested effects or other key elements.
(3) \textbf{Omission in Reasoning}  For errors where critical pathway steps in the question's biological process are left out, causing the final answer to be incorrect.
(4) \textbf{Faulty in Reasoning} When the reasoning path is correct, but there are significant errors in deducing the events within that pathway.

In this subsection, we provide examples of each failure reason type. The 'Question' and 'Answer' represent the input question and its labeled correct answer. The 'Wrong Answer by Model' refers to the incorrect response generated by the model. Additionally, we include a 'Reference Reasoning Process', which outlines the reasoning process leading to the correct final answer by LLaMA3.1 405B. This helps readers understand how errors occur during the reasoning process.

For the error types 'Omission in Reasoning' and 'Faulty Reasoning,' we present cases for both CoT and PathSeeker individually. These examples demonstrate how LLM reasoning fails either on its own or when augmented with a pathway graph.

\textbf{Category 1: Unresolved Conclusion}

\begin{tcolorbox}[breakable]

"Question": "A large fraction of pediatric pre-B acute lymphoblastoid leukemias (ALL) consistently contain a t(1;19) chromosomal translocation, resulting in the production of a chimeric transcription factor containing the N-terminal transactivation domain of E2A fused to the C-terminal DNA-binding homeodomain of Pbx1. What is the effect of the E2A-Pbx1 fusion protein on the expression of the novel WNT gene, WNT-16, in pre-B ALL cells?",

"Answer": "The E2A-Pbx1 fusion protein activates the expression of WNT-16 in pre-B ALL cells.",

\begin{tikzpicture}
\draw[dashed] (0,0) -- (\linewidth,0); 
\end{tikzpicture}

"Wrong Answer by Model": "The E2A-Pbx1 fusion protein in pediatric pre-B acute lymphoblastic leukemias (ALL) with the t(1;19) chromosomal translocation is known to have oncogenic properties and can affect gene expression. In the case of the novel WNT gene, WNT-16, the E2A-Pbx1 fusion protein may have an effect on its expression in pre-B ALL cells. However, the specific effect of the fusion protein on WNT-16 expression would depend on the regulatory elements present in the WNT-16 gene and the transcriptional activity of the fusion protein. Further investigation would be needed to determine the exact effect of the E2A-Pbx1 fusion protein on WNT-16 expression in pre-B ALL cells.",

\begin{tikzpicture}
\draw[dashed] (0,0) -- (\linewidth,0); 
\end{tikzpicture}

"Reference Reasoning Process": "To determine the effect of the E2A-Pbx1 fusion protein on the expression of the novel WNT gene, WNT-16, in pre-B ALL cells, we can analyze the provided pathways for transcriptional regulation. The pathways indicate that both E2A (6929) and Pbx1 are transcription factors and that their fusion can lead to transcriptional activation of target genes.

Pathways 0 and 13 demonstrate that the E2A-Pbx1 fusion protein can cause the expression of WNT-16. The fusion of E2A and Pbx1 results in a chimeric transcription factor that retains the transactivation domain of E2A and the DNA-binding homeodomain of Pbx1.

Considering this information, it can be inferred that the E2A-Pbx1 fusion protein leads to increased expression of the novel WNT gene, WNT-16, in pre-B ALL cells. The fusion of E2A and Pbx1 results in a transcription factor that activates the expression of WNT-16, potentially contributing to the development or progression of pre-B ALL.
\end{tcolorbox}

\begin{figure}[!h]
    \vspace{0.01cm}
    \caption{The Unresolved Conclusion error occurs when the model cannot provide a definitive answer, indicating uncertainty or the belief that the answer is unknown. In this case, the model is unable to determine the specific effect of the fusion protein on WNT-16 expression.}
\end{figure}

\textbf{Category 2: Incomplete Answer}
\begin{tcolorbox}[breakable]
"Question": "Continual efferocytic clearance of apoptotic cells (ACs) by macrophages is crucial for injury resolution. The process of continual efferocytosis is enhanced by linking the metabolism of engulfed cargo from initial efferocytic events to subsequent rounds. How is continual efferocytosis enhanced in macrophages?",

"Answer": "Continual efferocytosis is enhanced in macrophages by the metabolism of apoptotic cell-derived arginine and ornithine to putrescine by macrophage arginase 1 (Arg1) and ornithine decarboxylase (ODC).",

\begin{tikzpicture}
\draw[dashed] (0,0) -- (\linewidth,0); 
\end{tikzpicture}

"Wrong Answer by Model": "Continual efferocytosis is enhanced in macrophages through the process of metabolic reprogramming. When macrophages engulf apoptotic cells (ACs), they not only clear the ACs but also metabolize the engulfed cargo. This metabolic reprogramming involves the activation of specific metabolic pathways, such as glycolysis and fatty acid oxidation, to generate energy and metabolites necessary for the efficient clearance of ACs. By linking the metabolism of engulfed cargo from initial efferocytic events to subsequent rounds, macrophages can sustain their energy requirements and maintain the high phagocytic capacity needed for continual efferocytosis. This enhanced metabolic activity allows macrophages to efficiently clear ACs and promote injury resolution.",

\begin{tikzpicture}
\draw[dashed] (0,0) -- (\linewidth,0); 
\end{tikzpicture}

"Reference Reasoning Process": "To determine how continual efferocytosis is enhanced in macrophages, we can analyze the provided pathways for the key steps involved in efferocytic clearance of apoptotic cells (ACs) and how they link to subsequent rounds.

The process starts with the engulfment of apoptotic cells. This involves the recognition of phosphatidylserine on the surface of apoptotic cells by receptors such as AGER (Pathway 0). This recognition leads to the activation of signaling pathways, including the activation of Rac family small GTPase 1 (RAC1), which is crucial for the engulfment process (Pathway 1).

Following engulfment, the metabolism of engulfed cargo plays a critical role in enhancing continual efferocytosis. Specifically, the pathway involving the solute carrier family 66 member 1 (SLC66A1) and the metabolism of L-Arginine (Pathway 6) leads to the production of ornithine. Ornithine then binds to ARG1 (Pathway 7), leading to the production of urea and more ornithine (Pathway 8). This process enhances continual efferocytosis by providing the necessary components for subsequent rounds of efferocytic clearance.

Moreover, the pathway involving the metabolism of ornithine to putrescine (Pathway 9) and then the activation of RAC1 by putrescine (Pathway 11) further supports the enhancement of continual efferocytosis. This is because RAC1 activation is crucial for the engulfment process, and its activation by putrescine ensures that the macrophage remains capable of engulfing subsequent apoptotic cells.

Therefore, based on the pathways provided, continual efferocytosis in macrophages is enhanced by the metabolism of engulfed cargo, specifically through the arginine and ornithine pathways, which ultimately lead to the activation of key signaling molecules such as RAC1 that are necessary for the efferocytic clearance of apoptotic cells.",
\end{tcolorbox}
\begin{figure}[!h]
    \vspace{0.01cm}
    \caption{The Incomplete Answer error occurs when the response lacks essential details, such as missing the requested effects or other key elements. In this case, the model's response suggests that continual efferocytosis is enhanced in macrophages through metabolic reprogramming but fails to mention the metabolism of apoptotic cell-derived arginine and ornithine to putrescine, which is a key part of the labeled answer.}
\end{figure}

\textbf{Category 3: Omission in Reasoning}

\textbf{Omission in Reasoning of CoT}

\begin{tcolorbox}[breakable]
"Question": "BAMBI (BMP and activin membrane-bound inhibitor) is known to negatively regulate the signaling activity of transforming growth factor-beta, activin, and BMP, and has been identified as a target of beta-catenin in colorectal and hepatocellular tumor cells. The canonical Wnt/beta-catenin pathway plays a crucial role in regulating embryogenesis and tumorigenesis by promoting cell proliferation. Does BAMBI enhance or inhibit Wnt-promoted cell cycle progression?",

"Answer": "BAMBI increases Wnt-promoted cell cycle progression.",

\begin{tikzpicture}
\draw[dashed] (0,0) -- (\linewidth,0); 
\end{tikzpicture}

"Wrong Answer by Model": "BAMBI inhibits Wnt-promoted cell cycle progression. The canonical Wnt/beta-catenin pathway promotes cell proliferation, and BAMBI has been identified as a target of beta-catenin. BAMBI negatively regulates the signaling activity of transforming growth factor-beta, activin, and BMP, which are involved in cell cycle regulation. Therefore, BAMBI acts as a negative regulator of the Wnt pathway and inhibits Wnt-promoted cell cycle progression.",

\begin{tikzpicture}
\draw[dashed] (0,0) -- (\linewidth,0); 
\end{tikzpicture}

"Reference Reasoning Process": "To determine whether BAMBI enhances or inhibits Wnt-promoted cell cycle progression, we need to analyze its interactions with key components of the Wnt signaling pathway.

From pathways 10, 11, and 12, we can see that BAMBI (BMP and activin membrane-bound inhibitor) interacts with LRP6, FZD10, and DVL1, which are all crucial components of the Wnt signaling pathway. These interactions are described as "activation" and "binding/association," which suggests that BAMBI may be enhancing the activity of these components.

However, it's also known that BAMBI is a negative regulator of the TGF-beta and BMP signaling pathways, which can intersect with the Wnt signaling pathway. 

Considering the complexity of these interactions and the multiple pathways involved, we need to look for more direct evidence of BAMBI's effect on Wnt-promoted cell cycle progression.

Given the information provided, it seems that BAMBI's interactions with the Wnt signaling pathway are more likely to be enhancing or modulating rather than inhibiting. However, without more direct evidence, it's difficult to make a definitive conclusion.

Therefore, based on the available information, it appears that BAMBI may enhance Wnt-promoted cell cycle progression, but further research is needed to confirm this conclusion.",
\end{tcolorbox}
\begin{figure}[!h]
    \vspace{0.01cm}
    \caption{Omission in Reasoning refers to errors where critical steps in the biological pathway described in the question are omitted, leading to an incorrect final answer. In this case, the model's reasoning process only considered that BAMBI has been identified as a target of beta-catenin but overlooked the pathway in which BAMBI interacts with LRP6, FZD10, and DVL1, all of which are crucial components of the Wnt signaling pathway. As a result, the reasoning led to an incorrect conclusion.}
\end{figure}

\textbf{Omission in Reasoning of PathSeeker}

\begin{tcolorbox}[breakable]
"Question": "Angiotensin II (Ang II) is an octapeptide that plays a crucial role in the cardiovascular system, and its formation and degradation are tightly regulated. To investigate the role of mast cell proteases in Ang II production, ex vivo cultures of peritoneal cells from mouse strains lacking mast cell heparin or mouse mast cell protease 4 (mMCP-4) were used. What is the effect of heparin deficiency on the formation and degradation of Ang II in these peritoneal cell cultures?",

"Answer": "The formation and degradation of Ang II are strongly reduced in heparin-deficient peritoneal cells.",

\begin{tikzpicture}
\draw[dashed] (0,0) -- (\linewidth,0); 
\end{tikzpicture}

"Wrong Answer by Model": "Based on the provided pathways, the effect of heparin deficiency on the formation and degradation of Ang II in peritoneal cell cultures can be inferred. Heparin is known to play a role in the regulation of Ang II production.

Pathway 0 indicates that carboxypeptidase A3 (CPA3) is involved in the degradation of Ang II. This suggests that heparin deficiency may lead to a decrease in the degradation of Ang II, potentially resulting in increased levels of Ang II.

Pathway 1 shows that angiotensin I converting enzyme (ACE) is responsible for the formation of Ang II. However, this pathway does not directly provide information about the effect of heparin deficiency on Ang II formation.

Considering the information from pathways 0 and 1, it can be inferred that heparin deficiency may primarily affect the degradation of Ang II rather than its formation. Without heparin, the degradation of Ang II by carboxypeptidase A3 may be impaired, leading to increased levels of Ang II in the peritoneal cell cultures.",

\begin{tikzpicture}
\draw[dashed] (0,0) -- (\linewidth,0); 
\end{tikzpicture}

"Reference Reasoning Process": "To determine the effect of heparin deficiency on the formation and degradation of Ang II in peritoneal cell cultures, we can analyze the provided pathways for the role of mast cell proteases in Ang II production.

The pathways provided do not directly mention heparin or its deficiency. However, we can infer the potential effects of heparin deficiency by understanding the role of mast cell proteases in Ang II production.

Pathway 1 shows the conversion of Angiotensin I to Angiotensin III. Pathway 2 demonstrates the binding of Angiotensin I to angiotensin I converting enzyme (ACE), which is involved in the conversion of Angiotensin I to Angiotensin II. Pathway 3 shows the activation of angiotensin II receptor type 2 by Angiotensin II. Pathway 4 demonstrates the activation of AGTR1 and AGTR2 receptors by Angiotensin II. Pathway 5 shows the activation of angiotensin II receptor type 2 by Angiotensin A.

Considering the information provided by these pathways, we can infer that heparin deficiency may affect the activity of mast cell proteases, such as mouse mast cell protease 4 (mMCP-4), which are involved in the conversion of Angiotensin I to Angiotensin II. Heparin is a cofactor for the activation of mast cell proteases, and its deficiency may lead to reduced activity of these proteases.

Therefore, the effect of heparin deficiency on the formation and degradation of Ang II in peritoneal cell cultures may be a reduction in the conversion of Angiotensin I to Angiotensin II, leading to decreased levels of Angiotensin II.",
\end{tcolorbox}
\begin{figure}[!h]
    \vspace{0.01cm}
    \caption{Omission in Reasoning refers to errors where critical steps in the biological pathway involved in the question are omitted, leading to an incorrect final answer. In this case, the model's reasoning process considered the pathway involving the degradation of Ang II but missed the more important pathway of the conversion of Angiotensin I to Angiotensin II. This resulted in reasoning that led to an incorrect conclusion. }
\end{figure}

\textbf{Category 4: Faulty in Reasoning}

\textbf{Faulty in Reasoning of CoT}

\begin{tcolorbox}[breakable]
"Question": "Enteropathogenic Escherichia coli (EPEC) is a major causative agent of infant diarrhoea in developing countries, and the EspF effector protein is injected from EPEC into host cells via a type III secretion system, disrupting host intestinal barrier function and initiating the mitochondrial death pathway. To clarify the manner in which EspF affects host cells, eukaryotic EspF-binding proteins were identified using affinity purification, and Abcf2, a protein of unknown function and member of the ABC-transporter family, was found to bind EspF. What is the effect of knocking down Abcf2 expression on EspF-induced caspase cleavage?",

"Answer": "Knocking down Abcf2 expression increases EspF-induced caspase 9 and caspase 3 cleavage.",

\begin{tikzpicture}
\draw[dashed] (0,0) -- (\linewidth,0); 
\end{tikzpicture}

"Wrong Answer by Model": "The effect of knocking down Abcf2 expression on EspF-induced caspase cleavage is likely to decrease or inhibit caspase cleavage. Abcf2, as a binding protein of EspF, may play a role in the signaling pathway that leads to caspase activation and subsequent cleavage. By knocking down Abcf2 expression, the interaction between EspF and Abcf2 is disrupted, potentially preventing or reducing the activation of caspases and subsequent cleavage. This suggests that Abcf2 is involved in the EspF-induced mitochondrial death pathway and its knockdown may have a protective effect against the disruption of host intestinal barrier function caused by EPEC infection.",

\begin{tikzpicture}
\draw[dashed] (0,0) -- (\linewidth,0); 
\end{tikzpicture}

"Reference Reasoning Process": "To determine the effect of knocking down Abcf2 expression on EspF-induced caspase cleavage, we should analyze the provided pathways for the interactions between EspF, Abcf2, and caspase.

Pathway 0 provides information that EspF binds Abcf2, and this interaction inhibits Abcf2. Additionally, pathway 2 reveals that Abcf2 inhibits caspase 9, which is involved in the intrinsic apoptotic pathway.

When EspF binds Abcf2, it inhibits Abcf2's function. Since Abcf2 normally inhibits caspase 9, the inhibition of Abcf2 by EspF would lead to an increase in caspase 9 activity.

If Abcf2 expression is knocked down, its inhibitory effect on caspase 9 would be reduced. As a result, caspase 9 activity would increase, similar to the effect of EspF binding Abcf2.

Therefore, knocking down Abcf2 expression would enhance EspF-induced caspase cleavage. The reduction in Abcf2's inhibitory effect on caspase 9 would amplify the pro-apoptotic signal, leading to increased caspase cleavage.

In summary, the effect of knocking down Abcf2 expression on EspF-induced caspase cleavage would be an increase in caspase cleavage due to the reduced inhibitory effect of Abcf2 on caspase 9.",
\end{tcolorbox}
\begin{figure}[!h]
    \vspace{0.01cm}
    \caption{Faulty Reasoning occurs when the reasoning path is correct, but there are significant errors in deducing the events within that pathway. In this case, the model's reasoning step proposes that EspF binds to Abcf2 and activates it, which is incorrect, as it actually inhibits Abcf2. As a result, the reasoning leads to an incorrect conclusion.}
\end{figure}

\textbf{Faulty in Reasoning of PathSeeker}

\begin{tcolorbox}[breakable]
"Question": "Bacterial pathogens often manipulate host immune pathways to establish acute and chronic infection, and many Gram-negative bacteria do this by secreting effector proteins through a type III secretion system that alter the host response to the pathogen. In the case of Salmonella, the phage-encoded GogB effector protein targets the host SCF E3 type ubiquitin ligase through an interaction with Skp1 and the human F-box only 22 (FBXO22) protein. What is the effect of GogB-deficient Salmonella on tissue damage and colonization in the gut during long-term chronic infections?",

"Answer": "GogB-deficient Salmonella cause extensive tissue damage and enhanced colonization in the gut during long-term chronic infections.",

\begin{tikzpicture}
\draw[dashed] (0,0) -- (\linewidth,0); 
\end{tikzpicture}

"Wrong Answer by Model": "Based on the provided pathways, the effect of GogB-deficient Salmonella on tissue damage and colonization in the gut during long-term chronic infections can be inferred. 

Pathway 0 indicates that the GogB effector protein targets the host SCF E3 type ubiquitin ligase through an interaction with Skp1 and the human F-box only 22 (FBXO22) protein. This pathway is associated with Salmonella infection.

Pathways 1 and 2 provide information about other type III secretion system effectors (SseK1 and SseK3) and their interactions with TNF-NFKB signaling pathways. These pathways are also linked to Salmonella infection.

Pathway 3 shows the involvement of tumor necrosis factor (TNF) and its receptor TNFRSF1A in various signaling pathways, including Salmonella infection.

Pathway 4 indicates the activation of NFKB, a transcription factor involved in inflammation and immune responses, by TNF. This pathway is associated with Salmonella infection.

Pathways 5 and 6 provide information about the activation and inhibition of NFKB inhibitor alpha (NFKBIA) and its interaction with NFKB. These pathways are also linked to Salmonella infection.

Pathway 7 shows the activation of MAPK8 (JNK) and its phosphorylation by FOS, which is associated with Salmonella infection.

Based on the provided pathways, it can be inferred that GogB-deficient Salmonella would have a reduced ability to manipulate host immune pathways and alter the host response to the pathogen. This could result in decreased tissue damage and colonization in the gut during long-term chronic infections. However, the specific effects on tissue damage and colonization would require further investigation.",

\begin{tikzpicture}
\draw[dashed] (0,0) -- (\linewidth,0); 
\end{tikzpicture}

"Reference Reasoning Process": "Based on the pathways provided, the effect of GogB-deficient Salmonella on tissue damage and colonization in the gut during long-term chronic infections can be inferred.

GogB is a phage-encoded effector protein that targets the host SCF E3 type ubiquitin ligase through an interaction with Skp1 and the human F-box only 22 (FBXO22) protein (Pathway 0). This interaction inhibits the NF-kappaB signaling pathway, which is a crucial pathway involved in inflammation and immune response.

In the absence of GogB, the inhibition of the NF-kappaB signaling pathway is lifted, leading to an increase in the production of pro-inflammatory cytokines, such as TNF-alpha (Pathway 3). This increase in inflammation can lead to tissue damage and exacerbate the symptoms of the infection.

Furthermore, the NF-kappaB signaling pathway is also involved in the regulation of apoptosis (programmed cell death). The inhibition of this pathway by GogB can lead to a decrease in apoptosis, allowing the infected cells to survive and potentially leading to persistent infection (Pathway 6).

Additionally, GogB-deficient Salmonella may also have altered interactions with other host proteins, such as SifA, PipB2, and SopD2, which are involved in regulating microtubule dynamics and transport (Pathways 7, 9, and 10). These altered interactions can affect the ability of Salmonella to colonize and persist in the gut.

Overall, the loss of GogB in Salmonella is likely to lead to increased tissue damage and inflammation during long-term chronic infections, as well as altered interactions with host proteins that can affect the ability of the bacteria to colonize and persist in the gut.

In conclusion, the effect of GogB-deficient Salmonella on tissue damage and colonization in the gut during long-term chronic infections is likely to be an increase in tissue damage and inflammation, as well as altered interactions with host proteins that can affect the ability of the bacteria to colonize and persist in the gut.",
\end{tcolorbox}
\begin{figure}[!h]
    \vspace{0.01cm}
    \caption{Faulty Reasoning occurs when the reasoning path is correct, but there are significant errors in deducing the events within that pathway. In this case, the model reasoned along the NF-kappaB signaling pathway but failed to deduce that a GogB deficiency leads to an increase in inflammation. Consequently, the final conclusion is incorrect.}
\end{figure}


\subsection{Uncertainty Measure}

\begin{figure*}[!t]
    \centering    
    \renewcommand{\thesubfigure}{} 
    \subfigure[]{\includegraphics[width=0.48\linewidth]{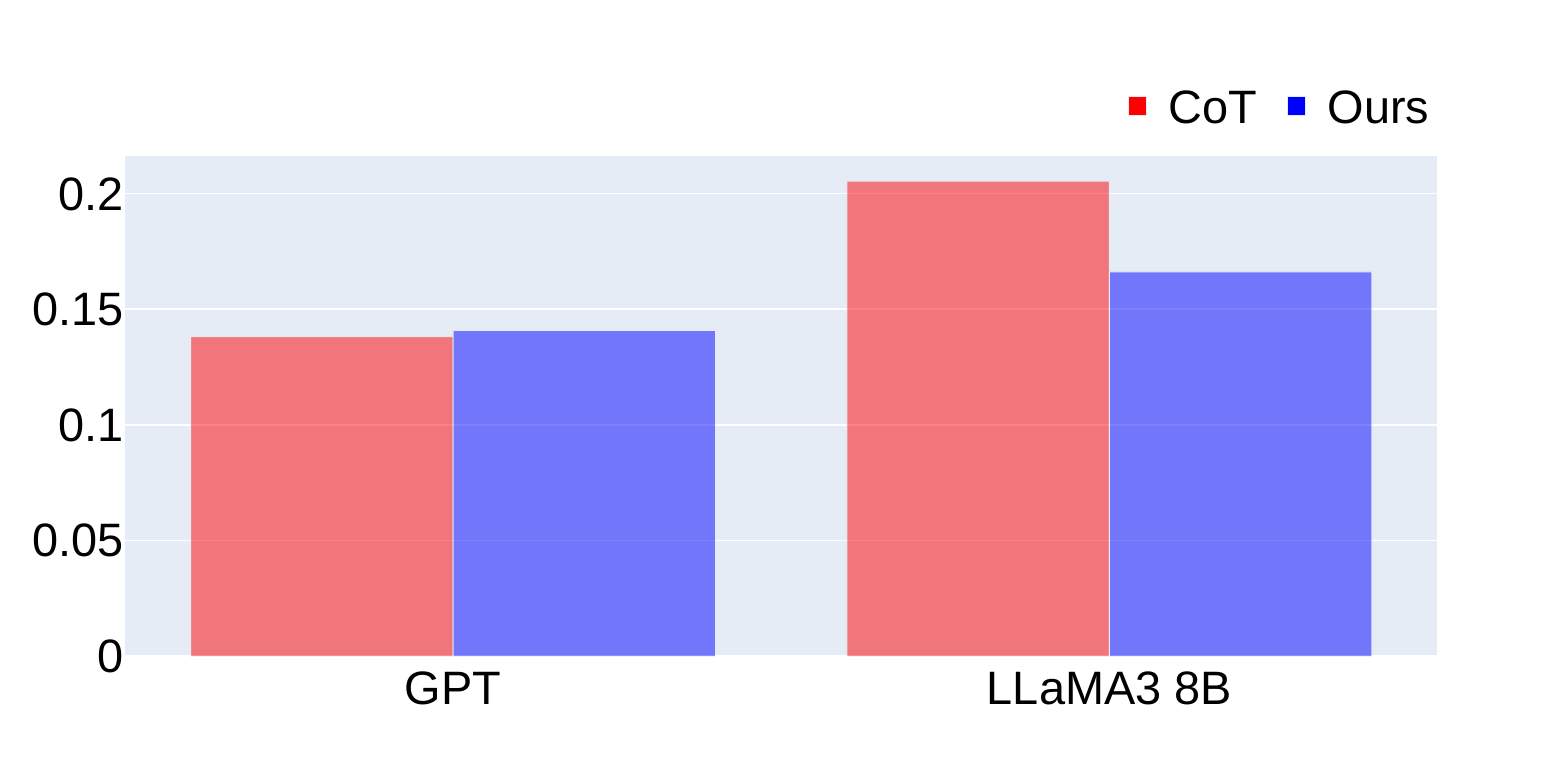}}
    \caption{Uncertainty measured by entropy.}
    \label{fig entropy}
\end{figure*}

\textbf{Uncertainty} We investigate whether graph augmentation can reduce the uncertainty in model responses. We measure this uncertainty in the discriminant task by calculating five times the entropy of the final results, as shown in Figure \ref{fig entropy}. Notably, graph augmentation reduces prediction uncertainty for LLaMA3-8B but not for GPT-3.5. This discrepancy may arise because GPT-3.5 tends to have fewer hallucinations, whereas LLaMA3-8B may exhibit overconfidence in some generations where it is uncertain.

\textbf{} \label{subsection evaluator quality}

\subsection{Evaluation of the Evaluator Quality}

As the generation task involves LLMs as evaluators, we assess the quality of the evaluation method by comparing the result with the human manual annotation score. The accuracy of LLaMA3.1 405B with human annotation is $96\%$, while the inconsistency lines in the case that the answer is close to the ground truth but the expression is general and missing details.

\subsection{Introduction of Pathway}

Understanding biological systems is inherently complex due to the numerous interacting molecules, processes, and environmental factors involved. These systems operate with intricate interactions that result in non-linear, multi-layered, and dynamic behaviors. To address this complexity, biological researchers use pathway graphs as structured blueprints to simplify these systems into organized structures that consist of basic interactions. The linear reactions, cyclical relationships, or the local network of pathways offer snapshots of how a system behaves under specific conditions and enable researchers to predict how changes in one molecule or interaction can affect the entire system. Pathway graphs also provide a structured, static representation of dynamic processes, helping researchers understand the sequence of events even as the system changes over time. 

\subsection{Detailed Related Work} \label{appendix related work}

\textbf{Biological Scientific Question Answering} Previous studies have explored the potential of language models in the biological scientific domain. MEDHOP \citep{welbl2018constructing} and PubMedQA \citep{jin2019pubmedqa} investigated biological scientific question answering in the form of reading comprehension. BioASQ-QA \citep{krithara2023bioasq} proposed a realistic question-answering benchmark for the actual information needs of biomedical experts. Beyond textual QA, several works have also studied multimodal scientific ability \citep{lu2022learn}. Additionally, other studies have explored biomedical domain tasks \citep{vilares2019head, jin2021disease, pal2022medmcqa}. Most existing tasks in the biological sciences concentrate on knowledge probing, assessing how well models understand biological information. In contrast, our work, \benchname, is the first to focus on models' reasoning abilities within the biological scientific domain, specifically targeting phenomena observed in experiments about biological pathways.

A few studies have examined language models' ability to understand biological pathways. Chatpathway \citep{li2023chatpathway} and \cite{azam2024comprehensive} specifically investigated language models' capacity for completing biological pathways. However, these studies mainly focus on probing biological pathway knowledge, i.e., determining whether language models possess the relevant pathway information. In contrast, this work introduces a novel task that employs pathway models for practical biological pathway phenomenon reasoning, bridging the gap between pathway network knowledge and its implications. We compare this work with previous biological datasets in Table \ref{appendix related work dataset table}.

\begin{table}[!h]
\caption{Comparison of previous biological scientific question answering tasks and \benchname.}
\label{appendix related work dataset table}
\resizebox{\linewidth}{!}{
\begin{tabular}{llll}
\toprule
& Domain                             & Question Form      &  Task Types                \\
\midrule
MEDHOP \cite{welbl2018constructing} & Scientific                         & Choice             & Multi-hop reading comprehension \\
PubMedQA \cite{jin2019pubmedqa}     & Scientific                         & True/False              &    Reading comprehension                                \\
HEAD-QA \cite{vilares2019head}      & {\color[HTML]{1F2328} Healthcare}  & Choice             & Knowledge probing and reasoning \\

MedQA \cite{jin2021disease}         & Medical                            & Choice             & Reading comprehension           \\
MedMCQA \cite{pal2022medmcqa}       & Medical                            & Choice             & Knowledge probing and reasoning \\

BioASQ-QA \cite{krithara2023bioasq} & Scientific & True/False and Open-Ended & Knowledege probing \\

ChatPathway \cite{li2023chatpathway}            & Pathway and biochemical & Open-Ended         & Knowledge probing               \\
\cite{azam2024comprehensive}        & Pathway and gene                   & Choice             & Knowledge probing               \\
\benchname (Ours)                         & Pathway for the scientific & True/False and Open-Ended & Reasoning in biological pathway        \\
\bottomrule
\end{tabular}
}
\end{table}

\textbf{Graph-augmented Language Model}
Several studies have explored augmenting large language models (LLMs) with graph data. In particular, some works enhance LLMs by encoding graph data as text \citep{ye2023natural, wang2024can, fatemi2023talk}, or tuning LLMs specifically for graph-based tasks \citep{liu2023one, tang2024graphgpt, he2024g, zhao2023gimlet, he2024unigraph}. By augmenting LLMs with graph data, they have been applied to knowledge-based QA \citep{sun2023think, he2024g, li2023chain, jin2024graph, cheng2024call}, and to graph-oriented tasks like graph property prediction \citep{wang2024can, he2023harnessing}. A few other studies leverage graph structures during LLM reasoning to tackle complex tasks \citep{jiang2023structgpt, besta2024graph}.

Unlike tasks in previous works, this study addresses whether reasoning in biological systems can be enhanced by pathway graphs, which act as a \textit{structured blueprint} for reasoning about the system’s states. It is not sufficient to simply identify the correct paths in the pathway graph to find the answer. Instead, it is necessary to perform deductive reasoning about the events that occur when the system is intervened upon under specific conditions and to predict the resulting states and mechanisms of the intervened system.

For large graph databases, most works enable LLMs to access graph data through retrieval mechanisms \citep{he2024g, li2023chain}, while a few studies have explored using LLMs as interactive agents \citep{yao2023react, shinn2023reflexion, zhao2024empowering} to navigate and explore vast graph databases \citep{sun2023think, jin2024graph}. In this work, we introduce an agent-based interactive graph exploration approach using subgraph navigation-based browsing, which is more efficient and offers enhanced navigation capabilities for pathway database.

\end{document}